\documentclass[sigconf]{acmart}

\usepackage{amsmath}
\newtheorem{theorem}{Theorem}
\newtheorem{lemma}{Lemma}
\newtheorem{assumption}{Assumption}

\newtheorem{definition}{Definition}

\usepackage{multirow}
\usepackage{mathrsfs}
\usepackage{graphics}
\usepackage{subfigure}
\usepackage{wrapfig}
\usepackage{epstopdf}
\usepackage{pdfpages}
\usepackage{diagbox}
\usepackage{ulem}
\usepackage{CJKulem}
\usepackage{tablefootnote}
\usepackage{tabularx}

\usepackage{makecell}
\usepackage{float}

\AtBeginDocument{%
  \providecommand\BibTeX{{%
    \normalfont B\kern-0.5em{\scshape i\kern-0.25em b}\kern-0.8em\TeX}}}

\copyrightyear{2023}
\acmYear{2023}
\setcopyright{acmlicensed}
\acmConference[KDD '23] {Proceedings of the 29th ACM SIGKDD Conference on Knowledge Discovery and Data Mining}{August 6--10, 2023}{Long Beach, CA, USA.}
\acmBooktitle{Proceedings of the 29th ACM SIGKDD Conference on Knowledge Discovery and Data Mining (KDD '23), August 6--10, 2023, Long Beach, CA, USA}
\acmPrice{15.00}
\acmISBN{979-8-4007-0103-0/23/08}
\acmDOI{10.1145/3580305.3599299}

\begin{document}

\title{Decoupled Rationalization with Asymmetric Learning Rates: A Flexible Lipschitz Restraint}

\author{Wei Liu}
\email{idc_lw@hust.edu.cn}
\orcid{0000-0002-3871-9454}
\affiliation{%
  \institution{School of Computer Science and Technology, Huazhong University of Science and Technology, China}
  \country{}
}

\author{Jun Wang}
\authornote{Corresponding authors.}
\email{jwang@iwudao.tech}
\orcid{0000-0002-9515-076X}
\affiliation{%
  \institution{iWudao.tech, China}
  \country{}
}

\author{Haozhao Wang}
\authornotemark[1]
\email{hz_wang@hust.edu.cn}
\orcid{0000-0002-7591-5315}
\affiliation{%
  \institution{School of Computer Science and Technology, Huazhong University of Science and Technology, China}
  \country{}
}

\author{Ruixuan Li}
\authornotemark[1]
\email{rxli@hust.edu.cn}
\orcid{0000-0002-7791-5511}
\affiliation{%
  \institution{School of Computer Science and Technology, Huazhong University of Science and Technology, China}
  \country{}
}

\author{Yang Qiu}
\email{anders@hust.edu.cn}
\orcid{0000-0002-3564-0521}
\affiliation{%
  \institution{School of Computer Science and Technology, Huazhong University of Science and Technology, China}
  \country{}
}

\author{Yuankai Zhang}
\email{yuankai_zhang@hust.edu.cn}
\orcid{0000-0001-7911-0170}
\affiliation{%
  \institution{School of Computer Science and Technology, Huazhong University of Science and Technology, China}
  \country{}
}

\author{Jie Han}
\email{jiehan@hust.edu.cn}
\orcid{0000-0002-3029-2476}
\affiliation{%
  \institution{School of Computer Science and Technology, Huazhong University of Science and Technology, China}
  \country{}
}

\author{Yixiong Zou}
\email{yixiongz@hust.edu.cn}
\orcid{0000-0002-2125-9041}
\affiliation{%
  \institution{School of Computer Science and Technology, Huazhong University of Science and Technology, China}
  \country{}
}


\renewcommand{\shortauthors}{Wei Liu et al.}

\begin{abstract}
  A self-explaining rationalization model is generally constructed by a cooperative game where a generator selects the most human-intelligible pieces from the input text as rationales, followed by a predictor that makes predictions based on the selected rationales. 
However, such a cooperative game may incur the degeneration problem where the predictor overfits to the uninformative pieces generated by a not yet well-trained generator and in turn, leads the generator to converge to a sub-optimal model that tends to select senseless pieces. 
In this paper, we theoretically bridge degeneration with the predictor's Lipschitz continuity. 
Then, we empirically propose a simple but effective method named DR, which can naturally and flexibly restrain the Lipschitz constant of the predictor, to address the problem of degeneration.
The main idea of DR is to decouple the generator and predictor to allocate them with asymmetric learning rates.
A series of experiments conducted on two widely used benchmarks have verified the effectiveness of the proposed method.  Codes: \href{https://github.com/jugechengzi/Rationalization-DR}{https://github.com/jugechengzi/Rationalization-DR}.
\end{abstract}





\normalem
\maketitle

\section{Introduction}

\begin{figure}[h]
    \centering
    \includegraphics[width=0.9\columnwidth]{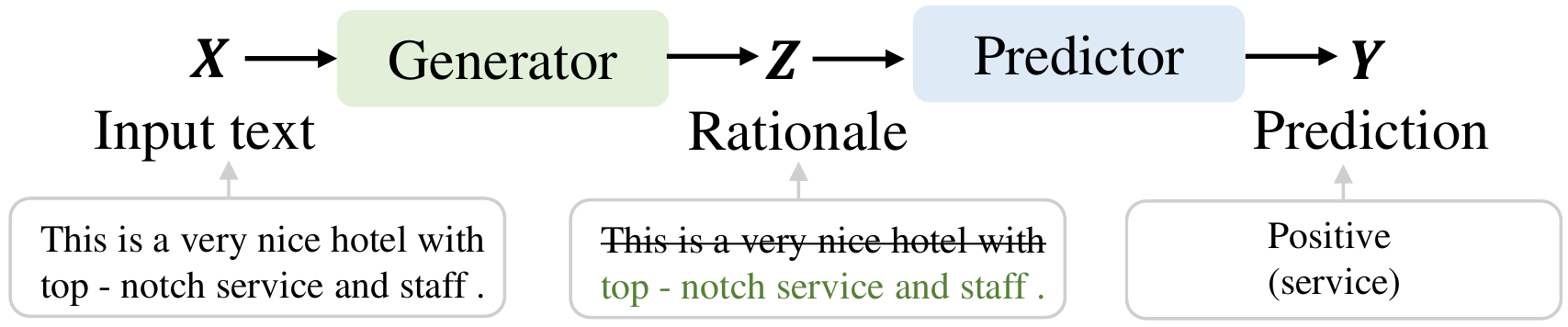}
    \caption{The standard rationalization framework RNP.   }
    \label{fig:RNP}
\end{figure}

The widespread use of deep learning in NLP models has led to increased concerns on interpretability.  \citet{emnlp/LeiBJ16} first proposed the rationalization framework RNP in which a generator selects human-intelligible subsets (i.e., rationales) from input text and feeds them to the subsequent predictor that maximizes the text classification accuracy, as shown in Figure~\ref{fig:RNP}. Unlike  post-hoc approaches for explaining blackbox models, the RNP framework has the built-in self-explaining ability through a cooperative game between the generator and the predictor. RNP and its variants have become the mainstream to facilitate the interpretability of NLP models \citep{rethinking,interlocking,dmr,liufr,mgr}.

\begin{table*}[t]
\caption{An example of RNP making the right sentiment prediction using the uninformative rationale. 
The \underline{underlined} piece of the text is the human-annotated rationale. The piece of the text in \textcolor{red}{red} is the rationale generated by RNP\protect\footnotemark. 
Initially, the generator may randomly select some uninformative candidates like “-” as rationales for the negative text. 
The predictor of RNP overfits to these uninformative rationales and classifies the sentiment according to whether “-” is included in the rationale. 
Guided by such a spoiled predictor, the generator in turn tends to select these uninformative rationales.}
\begin{tabularx}{\textwidth}{X}
\hline\hline
	\toprule
	
    \textbf{Label(Aroma):} Negative\\
    \textbf{Input text:}
         12 oz bottle poured into a pint glass - a - pours a transparent , pale golden color . the head is pale white with no cream , one finger 's height , and abysmal retention . i looked away for a few seconds and the head was gone \underline{s \textcolor{red}{\textbf{-}} {stale cereal grains dominate . hardly any}} \underline{{other notes to speak of . very mild} in strength} t - sharp corn/grainy notes throughout it 's entirety . watery , and has no hops characters or esters to be found. very simple ( not surprisingly ) m - highly carbonated and crisp on the front with a smooth finish d - yes , it is drinkable , but there are certainly better choices , even in the cheap american adjunct beer category . hell , at least drink original coors  \\
    \textbf{Rationale from RNP:} [“-”]\ \  
    \textbf{Prediction:} Negative\\
    \hline
    \hline
\end{tabularx}

\label{tab:egdegeneration}
\end{table*}

However, the generator and predictor in the vanilla RNP often are not well coordinated, and there are some major stability and robustness issues named as \textbf{degeneration} \citep{rethinking}.  As illustrated in Table~\ref{tab:egdegeneration}, the predictor may overfit to meaningless rationale candidates selected by the not yet well-trained generator, leading the generator to converge to the sub-optimal model that tends to output these uninformative candidates as rationales. To address this problem, most current methods introduce supplementary modules to regularize the generator \cite{rethinking} or the predictor \citep{dmr,interlocking,liufr}. 
But they all treat the two players equally and endow them with the same learning strategies, in which the spoiled predictor can only be partially calibrated by the extra modules. The underlying coordination mechanism between two players, which is one of the essential problems in multi-agent systems \citep{AAAI20sterberg}, has not been fully explored in rationalization.%

Lipschitz continuity is a useful indicator of model stability and robustness for various tasks, such as analyzing robustness against adversarial examples~\citep{Szegedy-iclr2014,nips18Lipschitz,iclr18evaluate}, convergence stability of Discriminator in GANs~\citep{arjovsky-icml2017,zhou19c-icml-2019} and stability of closed-loop systems with reinforcement learning controllers \citep{nips19Lipschitz}. 
Lipschitz continuity reflects surface smoothness of functions corresponding to prediction models, and is measured by Lipschitz constant (Equation~\ref{eqa:lip_c}). Smaller Lipschitz constant represents better Lipschitz continuity.
For unstable models in optimization, their function surfaces usually have some non-smooth patterns leading to poor Lipschitz continuity, such as steep steps or spikes, where model outputs may make a large change when the input values change by only a small amount. 

\footnotetext{The example of RNP is from \citep{liufr}.}

In this paper, we make contributions as follows:

First, we theoretically link degeneration to the predictor's Lipschitz continuity and find that a small Lipschitz constant makes the predictor more robust and less susceptible to uninformative candidates, thus degeneration is less likely to occur. The theoretical results open new avenues for this line of research, which is the main contribution of this paper.

Second, different from existing methods, we propose a simple but effective method called \textbf{D}ecoupled \textbf{R}ationalization (DR), which decouples the generator and predictor in rationalization by making the learning rate of the predictor lower than that of the generator, and flexibly limits the Lipschitz constant of the predictor with respect to the selected rationales without manually chosen cutoffs, so as to address degeneration without any changes to the basic structure of RNP. 

Third, empirical results on two widely used benchmarks show that DR significantly improves the performance of the standard rationalization framework RNP without changing its structure and outperforms several recently published state-of-the-art methods by a large margin.

\section{Related Work}\label{sec:related}
\textbf{Rationalization}. The base cooperative framework of rationalization named RNP \citep{emnlp/LeiBJ16} is flexible and offers a unique advantage: certification of exclusion, which means any unselected input is guaranteed to have no contribution to prediction \citep{interlocking}. Based on this cooperative framework, many methods have been proposed to improve RNP from different aspects. One series of research efforts focus on refining the sampling  process in the generator from different perspectives so as to improve the rationalization. \citet{2018rationalegumble} used 
Gumbel-softmax to do the reparameterization for binarized selection. \citet{hardkuma}  replaced the Bernoulli sampling
distributions with rectified Kumaraswamy distributions. \citet{jain2020faith} disconnected the training
regimes of the generator and predictor networks using a saliency threshold. \citet{informationbottle} imposed a discrete bottleneck objective to balance the task performance and the rationale length. \citet{mgr} tried to address the problems of spurious correlations and degeneration simultaneously with multiple diverse generators. \citet{irrationality} call for more rigorous evaluations of rationalization models. \citet{scott}  leverage meta-learning techniques to improve the quality of the explanations. \citet{cooperative} cooperatively train the models with standard continuous and discrete optimisation schemes. Other methods like data augmentation with pretrained models \citep{counter}, training with human-annotated rationales \citep{Unirex} have also been tried. These methods are orthogonal to our research. Another series of efforts seek to regularize the predictor
using supplementary modules which have access to the information of the full text \citep{rethinking,dmr,interlocking} such that the generator and the predictor will not collude to uninformative rationales.
3PLAYER \citep{rethinking} takes the unselected text $Z^c$ into consideration by inputting it to a supplementary predictor \emph{Predictor}$^c$. DMR \citep{dmr} tries to align the distributions of rationale with the full input text in both the output space and feature space. A2R \citep{interlocking} endows the predictor with the information of full text by introducing a soft rationale. FR \cite{liufr} folds the two players to regularize each other by sharing a unified encoder.   These methods are most related to our work, but none of them consider the relationship of the two players in this rationalization cooperative game. The underlying mechanisms of degeneration have not been fully explored.

\textbf{Asymmetric learning rates in game theory}. Practices and theories that lead to asymmetric learning rates for different modules in a game have been studied \citep{TTUR0,TTUR}. Previous practices mainly aims to reach a Nash equilibrium in a general sum game. \citet{TTUR0} showed that a asymmetric learning rates in general-sum games ensures a stationary local Nash equilibrium if the critic learns faster than the actor. \citet{TTUR} argued that GANs \citep{gan} can converge to a stationary local Nash equilibrium by reducing the generator's learning rate. 
It is worthwhile to note that although our method looks similar to them, the core strategy is completely different. They analyze the efficiency of converging to a general Nash equilibrium in adversarial games. While in the cooperative game like rationalization, convergence is not a problem. Instead, we face a problem of equilibrium selection \cite{interlocking,AAAI20sterberg}. In fact, our approach is the opposite of theirs: they speed up the critic while we slow down it.

\section{Preliminaries}\label{sec:preliminaries}

\subsection{Cooperative rationalization}


We consider the classification problem, where the input is a text sequence  ${X}$=$[x_1,x_2,\cdots,x_l]$ with ${x}_i\in \mathcal{R}^d$ being the $d$-dimensional word embedding of the $i$-th token and $l$ being the number of tokens. The label of ${X}$ is a one-hot vector $Y\in\{0,1\}^c$, where $c$ is the number of categories. $\mathcal{D}$ represents the training set. $\mathcal{Z}$ represents the set of rationale candidates selected by the generator from $X\in \mathcal{D}$.

Cooperative rationalization consists of a generator $f_G(\cdot)$ and a predictor $f_P(\cdot)$, and $\theta_g$ and $\theta_p$ represent the parameters of the generator and predictor, respectively.
The goal of the generator is to select the most informative pieces from the input. 

For each sample $(X,Y)\in \mathcal{D}$, the generator firstly outputs a sequence of binary mask $M=[m_1,\cdots,m_l]\in \{0,1\}^l$. Then, it forms the rationale $Z$ by the element-wise product of $X$ and $M$:
\begin{equation}\label{eqa:getrat}
    Z=M\odot X=[m_1x_1,\cdots,m_lx_l].
\end{equation}
To simplify the notation, we denote $f_G(X)$ as $Z$ in the following sections, i.e., $f_G(X)=Z$. But note that the direct output of the generator is a sequence of independent Bernoulli distributions from which $M$ is sampled.
In cooperative rationalization, the generator and the predictor are usually optimized cooperatively by minimizing the cross-entropy:
\begin{equation}\label{eqa:objpg}
    \mathop{\min}\limits_{\theta_g,\theta_p}\sum_{(X,Y) \in \mathcal{D}}H(Y,f_P(f_G(X))).
\end{equation}

To make the selected rationales human-intelligible, the original RNP constrains the rationales by short and coherent regularization terms. In this paper, we use the constraints updated by \citet{car}:
\begin{equation}\label{eqa:regular}
\Omega (M) = \lambda_1 \bigg \lvert \frac{||M||_1}{l}-s \bigg\rvert +\lambda_2\sum_{t=2}^{l} \big|m_t-m_{t-1} \big|. 
\end{equation} The first term encourages that the percentage of the tokens being selected as rationales is close to a pre-defined level $s$. The second term encourages the rationales to be coherent.

\subsection{Lipschitz Continuity}

In this paper, we consider the Lipschitz continuity over the general distance metric $d\in \mathcal{M}$, where $\mathcal{M}$ denotes the space of distance metric and the formal definition of $\mathcal{M}$ is in Definition~\ref{def:properties of distance} of Appendix~\ref{app:distance}.
\begin{definition}\label{def:Lipschitz}
A function $f: \mathcal{R}^n\xrightarrow{}\mathcal{R}^1$ is Lipschitz continuous on $\mathcal{X}\subset \mathcal{R}^n$ if there exists a constant $L \geq 0$ such that 
\begin{equation}\label{eqa:lip}
   \forall{x_i,x_j}\in \mathcal{X},\  |f(x_i)-f(x_j)|\leq L\cdot d(x_i,x_j),
\end{equation}
over the distance metric $d\in \mathcal{M}$. 

\end{definition}
A widely used $d$ is the norm, i.e., $d(x_i,x_j)=||x_i-x_j||_p$, 
where $p$ is the order of the norm. In the following sections, we use the general $d$ for theoretical analysis and the concreted form for quantitative experiments.
The smallest $L$ is called the Lipschitz constant, denoted as 
\begin{equation}\label{eqa:lip_c}
    L_c=\mathop{\sup}_{x_i,x_j\in \mathcal{X}}\frac{|f(x_i)-f(x_j)|}{||x_i-x_j||_p},
\end{equation}
$L_c$ represents the maximum ratio between variations in the output and variations in the input of a model, and is used to measure Lipschitz Continuity. Note that the specific number depends on the definition of the distance metric $d$. Our theoretical analysis in Section~\ref{sec:lipcausedegeneration} does not involve a specific definition of $d$.

\section{Correlation between Degeneration and Predictor's Lipschitz Continuity}\label{sec:lipcausedegeneration}

\begin{figure}[t]

    \centering
        \includegraphics[width=5cm]{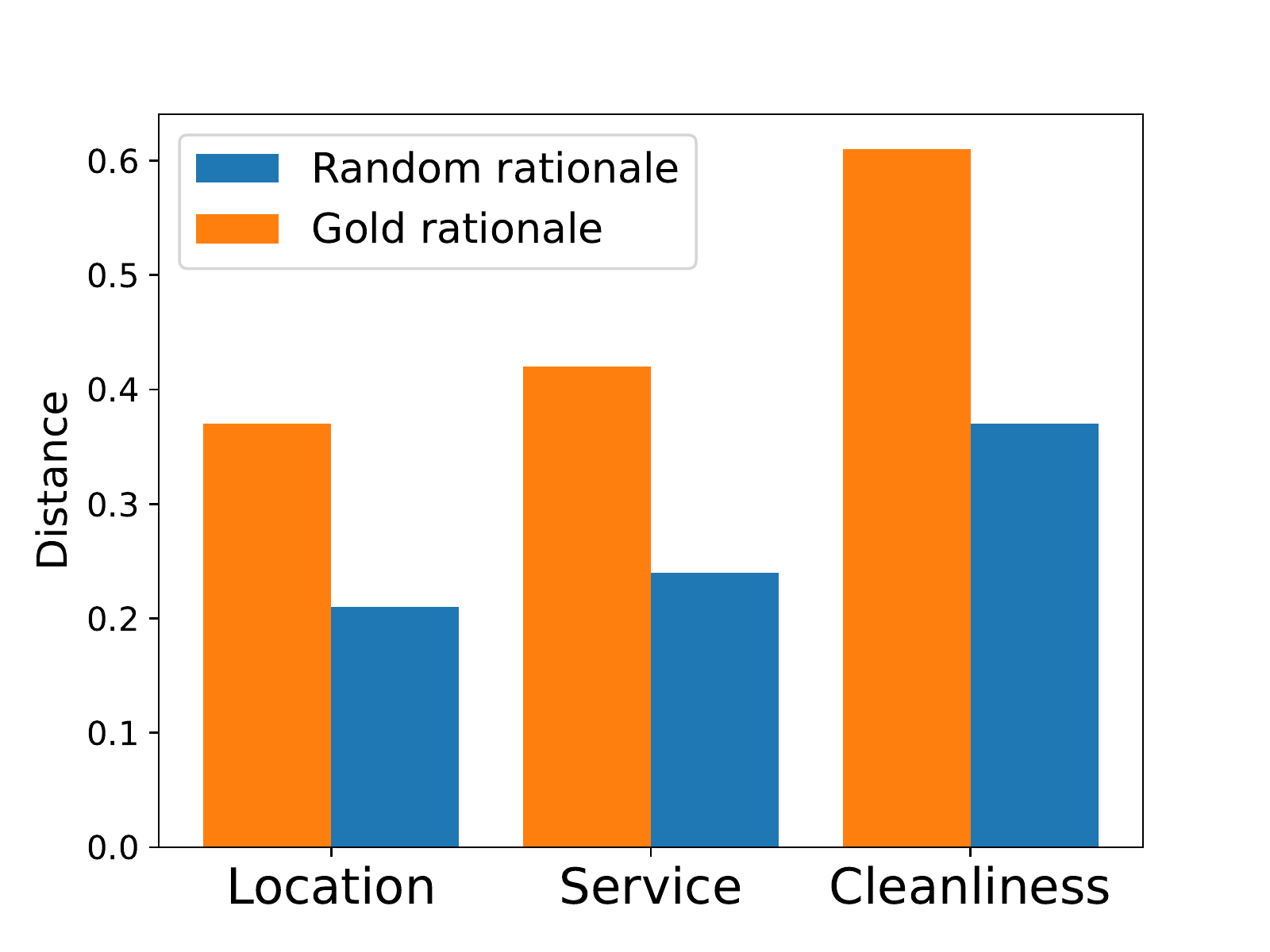}
        
  \caption{Average distance between rationales from positive reviews and rationales from negative reviews. Orange bars show average distance between human-annotated gold rationales from positive reviews and counterparts from negative reviews. Blue bars show average distance between randomly selected rationales from positive reviews and counterparts from negative reviews. The details are in Appendix~\ref{app:distance}.}
  \label{fig:distance}
\end{figure}

To demystify the correlation between degeneration and Lipschitz continuity of the predictor, without losing generality, we consider the binary classification problem such as sentiment analysis. 
Let $X_i$ and $X_j$ be two input texts with different classification labels: $Y_i=0$ (negative) and $Y_j=1$ (positive). $Z_i=f_G(X_i)$ and $Z_j=f_G(X_j)$ are two rationale candidates sampled from $X_i$ and $X_j$, respectively. We denote $\epsilon_i=f_P(Z_i)-0$ and $\epsilon_j=1-f_P(Z_j)$ as the discrepancies between their predictions and the true labels, and a smaller discrepancy indicates higher confidence of a prediction. $d(Z_i,Z_j)$ is the semantic distance between $Z_i$ and $Z_j$.

\subsection{Distance Properties of Rationales} 
Intuitively, if $Z_i$ and $Z_j$ are uninformative rationale candidates with no sentiment tendency, then their semantic distance $d(Z_i,Z_j)$ is generally small, and if $Z_i$ and $Z_j$ are informative rationales with clear sentiment tendency opposite to each other, then their semantic distance $d(Z_i,Z_j)$ is relatively larger. The above observation is supported by the practical experimental analyses in Figure~\ref{fig:distance}. The dataset is a multi-aspect classification dastset Hotel Reviews \citep{hotel}. Each aspect is calculated independently. More details of the experimental setup are in Appendix~\ref{app:distance}. The results show that if $Z_i$ and $Z_j$ are randomly selected (i.e., uninformative candidates), their average distance between each other is much smaller than those of human-annotated informative rationales under a specific distance metric defined in Appendix~\ref{app:distance}. 
For rigorous quantitative analysis, we quantify the qualitative intuition with an assumption:
\begin{assumption}\label{assumption:distance threshold}
    Given any two rationale candidates $Z_i$ and $Z_j$ that are selected from the inputting texts $X_i$ and $X_j$ with label $Y_i=0$ and $Y_j=1$, an arbitrary tolerable error probability $P_{te}$ and a distance metric $d$, there  exists a threshold $\delta_p$ corresponding to $P_{te}$ such that 
    \begin{equation}\label{eqa:assumption1}
        \text{if} \ d(Z_i,Z_j)\geq \delta_p,\  \text{then}\  \Tilde{p}(Z_i,Z_j)\leq P_{te} , 
    \end{equation}
     where $\Tilde{p}(Z_i,Z_j)$ denotes the probability that $Z_i$ and $Z_j$ are not both informative.
\end{assumption}

\subsection{Small Lipschitz Constant Leading to High Likelihood of Informative Candidates} 
We first qualitatively show that the generator is highly inclined to select informative rationales when the predictor is well trained and its Lipschitz constant is small, which is based on the intuition of the experiment of Figure~\ref{fig:distance} and does not involve Assumption~\ref{assumption:distance threshold}. Then, we formally quantify it in Theorem~\ref{the:rationale and lipschitz} based on Assumption~\ref{assumption:distance threshold}. 

Concretizing Equation~\ref{eqa:lip_c} into the predictor with 
the rationale candidate set $\mathcal{Z}$, and if the prediction error is small (i.e., $\epsilon_i,\epsilon_j\leq 0.5$), the corresponding Lipschitz constant of the
predictor is 
\begin{equation}\label{eqa:lip, distance and error}
\begin{aligned}
    L_c=&\mathop{\sup}_{Z_i,Z_j \in \mathcal{Z}}\frac{|f_P(Z_i)-f_P(Z_j)|}{d(Z_i,Z_j)}\\
    =&\mathop{\sup}_{Z_i,Z_j \in \mathcal{Z}}\frac{|(\epsilon_i+0)-(1-\epsilon_j)|}{d(Z_i,Z_j)}\\
    =&\mathop{\sup}_{Z_i,Z_j \in \mathcal{Z}}\frac{1-\epsilon_j-\epsilon_i}{d(Z_i,Z_j)}.
\end{aligned}
\end{equation}
$L_c$ is the supremum, so we have $L_c\geq \frac{1-\epsilon_j-\epsilon_i}{d(Z_i,Z_j)}$, and we further get 
\begin{equation}\label{eqa: distance larger than lip}
    d(Z_i,Z_j)\geq \frac{1-\epsilon_j-\epsilon_i}{L_c}.
\end{equation}
When the RNP model gives high-confidence predictions close to the true labels, $\frac{1-\epsilon_j-\epsilon_i}{L_c}$ can be seen as a lower bound of $d(Z_i,Z_j)$. 
Since informative candidates containing clear sentiment tendencies tend to have a greater distance between them than those uninformative ones, if $L_c$ is small enough, the predictor can make the right predictions (i.e., low $\epsilon_i,\epsilon_j$) with only informative candidates which have large $d(Z_i,Z_j)$, forcing the generator to select these candidates as rationales. 
Quantitatively, we have: 
\begin{theorem}\label{the:rationale and lipschitz}
Under Assumption~\ref{assumption:distance threshold}, given any two rationale candidates $Z_i$ and $Z_j$ that are selected from the inputting texts $X_i$ and $X_j$ with label $Y_i=0$ and $Y_j=1$, if the predictor makes correct predictions (i.e., $\epsilon_i,\epsilon_j\leq 0.5$) and its Lipschitz constant is small (i.e., $L_c\leq\frac{1-\epsilon_j-\epsilon_i}{\delta_p}$), then the probability that both $Z_i$ and $Z_j$ are informative rationales is at least $1-P_{te}$.
\end{theorem}
The proof is in Appendix~\ref{proof:rationale and lipschitz}.
As a consequence, constraining the Lipschitz constant of the predictor to be small can ease the degeneration problem where the predictor makes the right predictions with uninformative rationales.

\begin{figure}[t]
    \flushleft
    \subfigure[rationale quality]{
    \includegraphics[width=0.477\columnwidth]{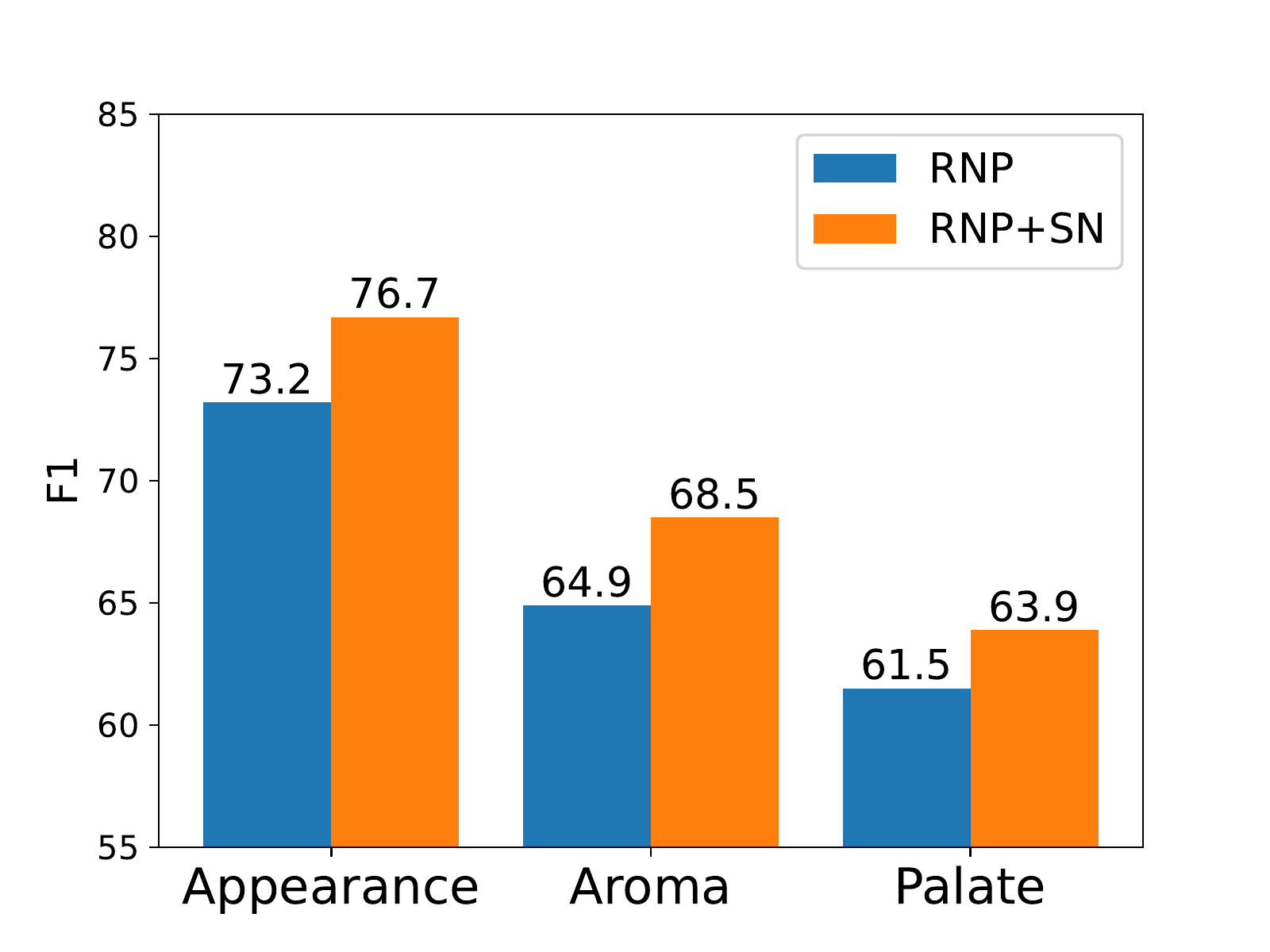}
    \label{fig:RNP_SN(a)}
        }
      \subfigure[prediction performance]{
        \includegraphics[width=0.477\columnwidth]{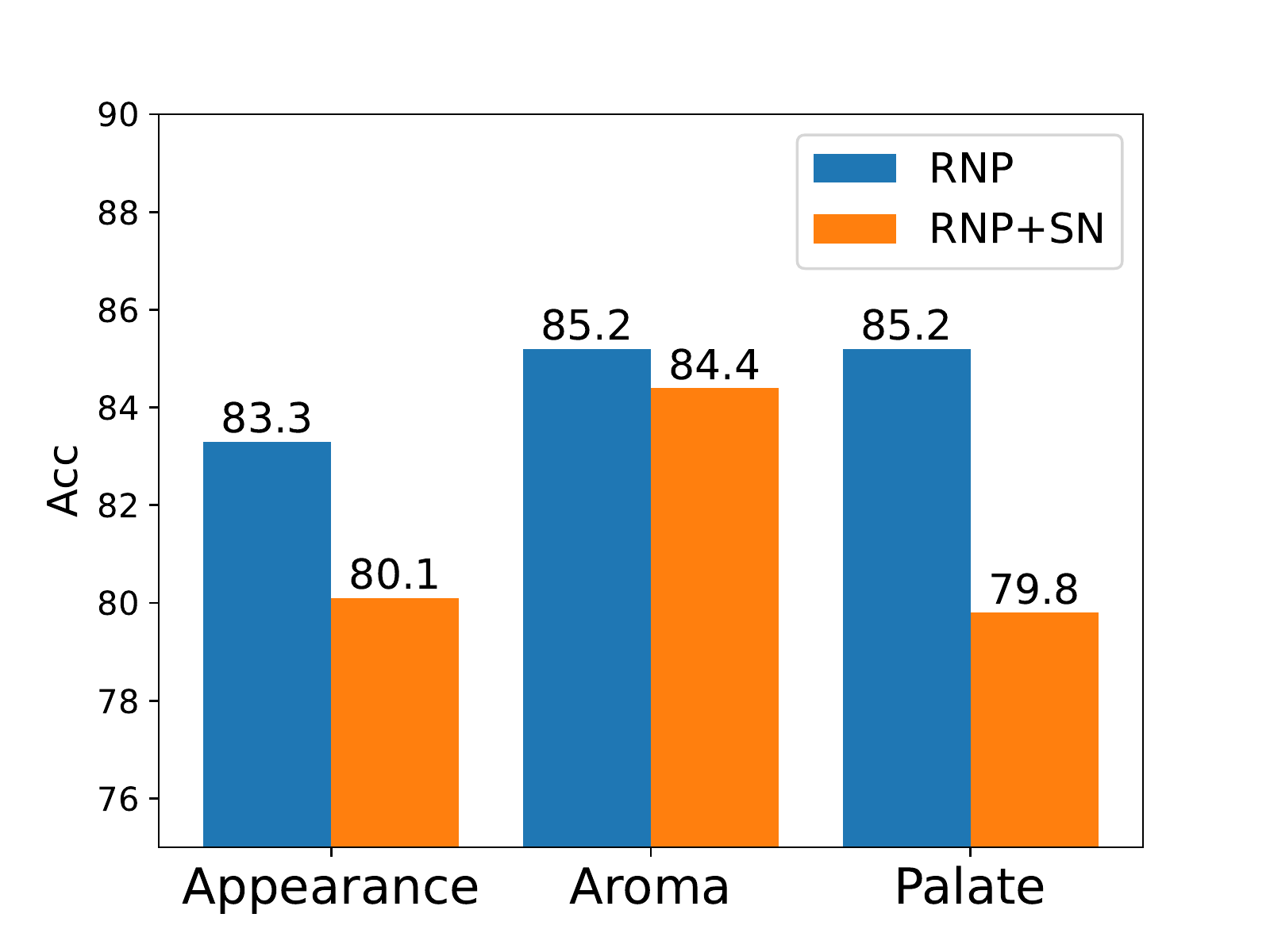}
        \label{fig:RNP_SN(b)}
        }

  \caption{An experiment on \emph{BeerAdvocate} restricting vanilla RNP's predictor's Lipchitz constant with spectral normalization. (a) and (b) report the rationale quality (F1) and prediction performance (Acc) respectively. $``$RNP+SN$"$: RNP with spectral normalization on predictor's linear layer.  }
  \label{fig:RNP+SN}

\end{figure}

\noindent\textbf{Verification with spectral normalization}.
There have been some existing methods such as weight clipping \citep{wgan} and spectral normalization \citep{spectral}, which can restrict the Lipschitz constant with some manually selected cutoff values. To verify the theoretical result that small Lipschitz constant improving the rationale quality, we conduct an experiment by simply applying the widely used spectral normalization \cite{spectral} to RNP's predictor's linear layer for restricting the Lipschitz constant. Figure~\ref{fig:RNP_SN(a)} shows that on all three independently trained datasets the rationale quality is significantly improved when the spectral normalization is applied, which demonstrates the effectiveness of restricting Lipschitz constant.

However, the manually selected cutoff values are not flexible and often limit the model capacity \citep{gradnorm,rethinking_lip}. In fact, different datasets need different values of the Lipschitz constant to maintain the model capacity in prediction. Figure~\ref{fig:RNP_SN(b)} shows that spectral normalization hurts the prediction performance.
Besides, these existing methods are also hard to be applied to complex network structures, such as RNN (including GRU and LSTM) and Transformer, which have multiple sets of parameters playing different roles that should not be treat equally. Gradient penalty \citep{WGAN-GP}, which is another method of restricting the Lipschitz constant used in GANs,  also fails because its objective function is not suitable for classification tasks and it is hard to create meaningful mixed data points due to the different lengths of input texts.

Instead of roughly truncating the Lipschitz constant, we use a simple method to guide the predictor to obtain an adaptively small Lipschitz constant without affecting its capacity, which can not only be applied to the RNN or Transformer based predictor in order to ease degeneration and improve rationale quality but also can maintain the good prediction performance.

\begin{figure}[t]
    \flushleft
      \subfigure[]{
        \includegraphics[width=0.477\columnwidth]{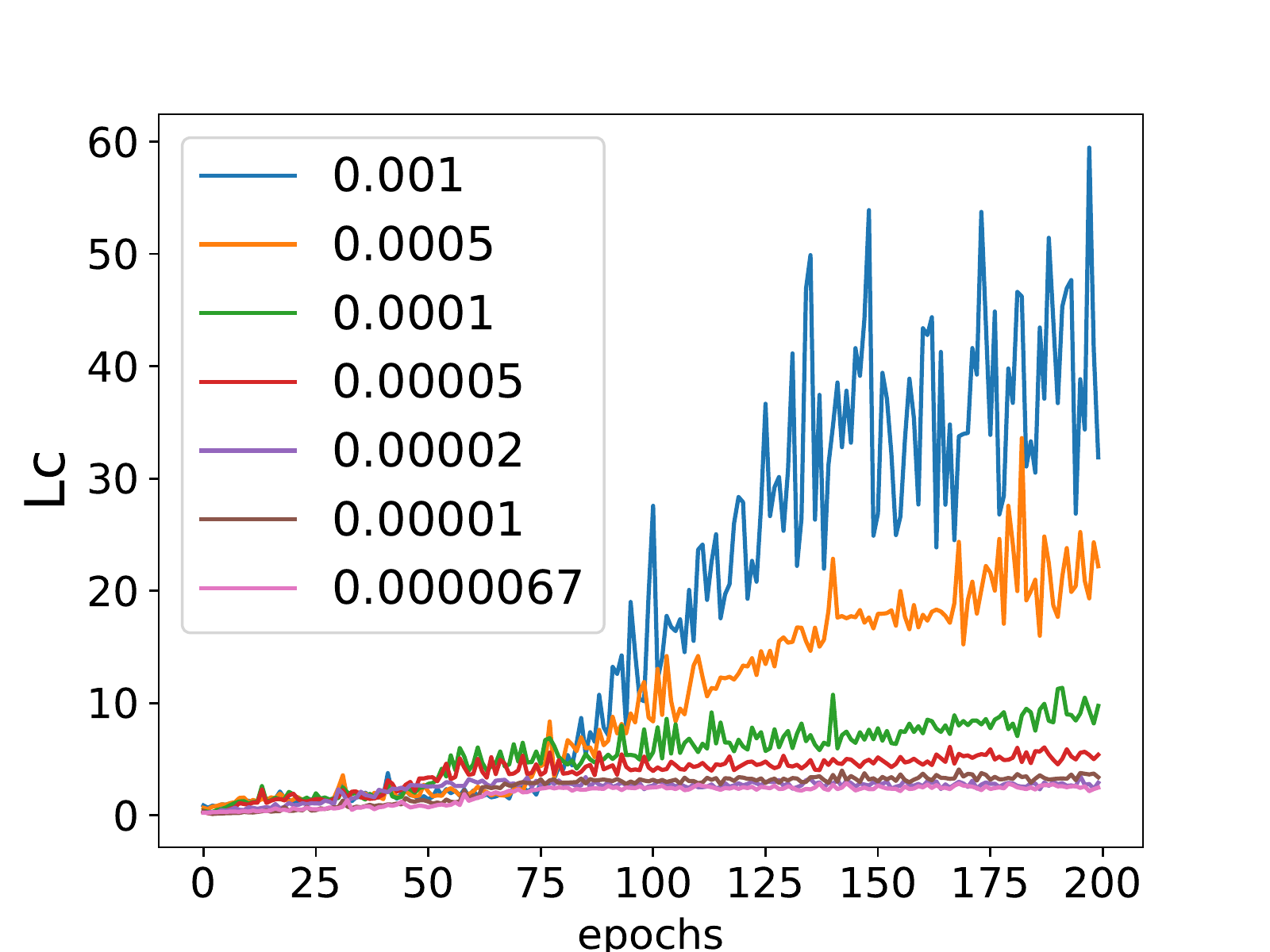}
        }
    \subfigure[]{
        \includegraphics[width=0.477\columnwidth]{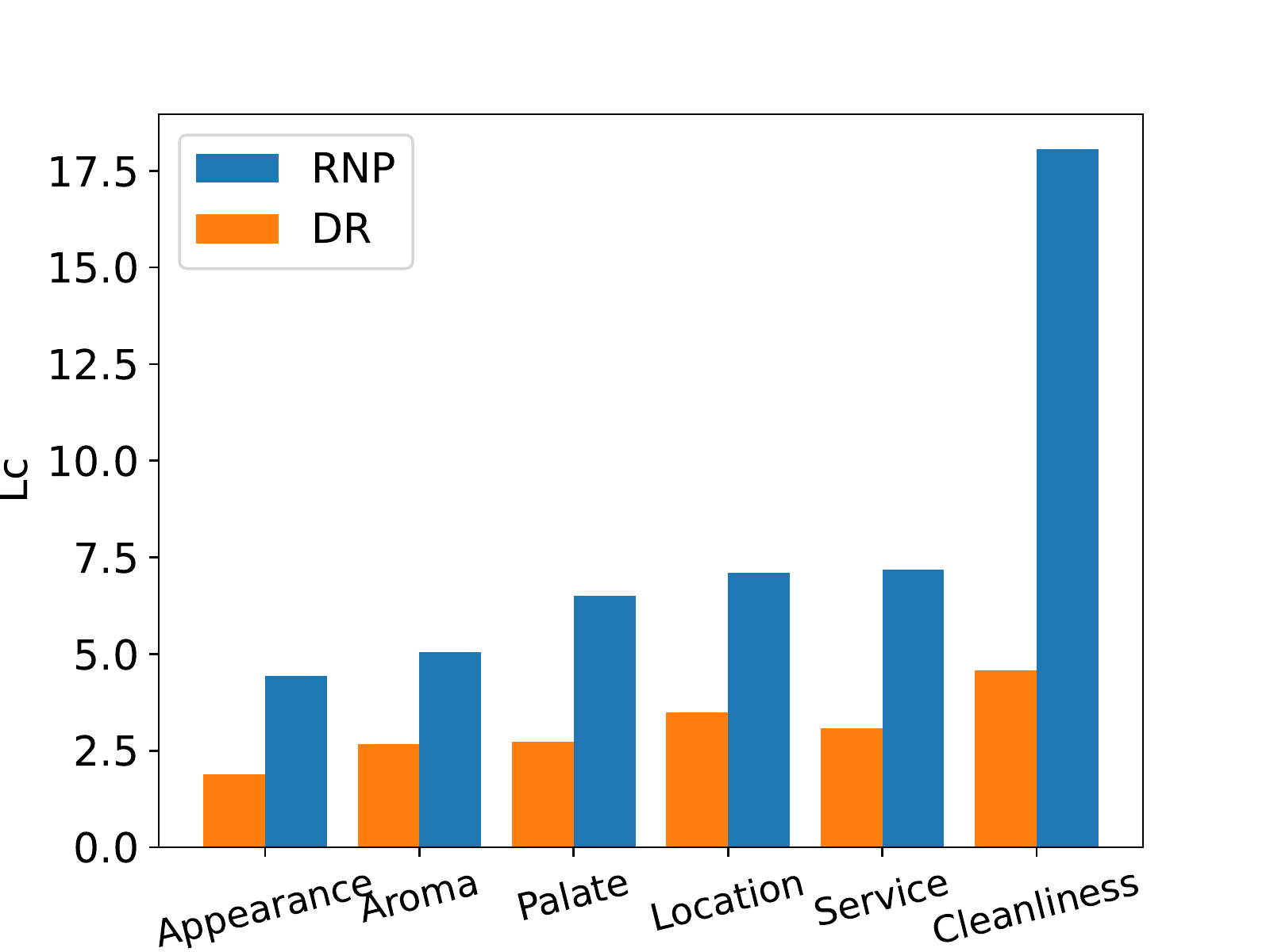}
    }

  \caption{
  (a): The approximate Lipschitz constant of the predictor with respect to selected rationales $Z$ using different predictor learning rates. Models are trained on \emph{Beer-Appearance} with the learning rate of the generator fixed at 0.0001.
  (b): Comparison of approximate Lipschitz constants between the \textcolor{blue}{RNP} and our \textcolor{orange}{DR} on different datasets. The details of the calculation of $L_c$ are in Appendix~\ref{app:cal_lc}.
  }
  \label{fig:lip with different pred lr}

\end{figure}

\section{Decoupled Rationalization with Asymmetric Learning Rates}

\subsection{Correlation Between Learning Rates and the Lipschitz Constant}\label{sec:relation on lip and lr}

\noindent\textbf{Qualitative analysis}.
Back to Equation~\ref{eqa: distance larger than lip}, we have 
\begin{equation}\label{eqa:error greater than 1-ld}
    \epsilon_i+\epsilon_j\geq 1-d(f_G(X_i),f_G(X_j))\cdot L_c ,
\end{equation}
where $L_c$ is related to $\theta_p$ and $d(f_G(X_i),f_G(X_j))$ is related to $\theta_g$. The goal of Equation~\ref{eqa:objpg} is to minimize the prediction discrepancies $\epsilon_i$ and $\epsilon_j$, which can be obtained by increasing the product $d(f_G(X_i),f_G(X_j))\cdot L_c$ by tuning $\theta_g$ and $\theta_p$. By tuning $\theta_g$, the generator tries to find rationales that get large distance between different categories so as to reduce the prediction discrepancies with greater $d(f_G(X_i),f_G(X_j))$. For the predictor, $L_c$ will be increased when it tunes $\theta_p$ to get lower $\epsilon$ on given rationale candidates, which is also reflected in Equation~\ref{eqa:lip, distance and error}. In short, both $d(f_G(X_i),f_G(X_j))$ and $L_c$ tends to grow during training,  but we want $L_c$ to be restricted to a small extent for better rationale quality.
 So, if we slow down the predictor and speed up the generator, chances are that the increase 
 in the product comes mainly from increasing $d(f_G(X_i),f_G(X_j))$ by selecting rationales with  clear sentiment tendencies rather than increasing $L_c$ by overfitting the predictor quickly.

\noindent\textbf{Empirical support}.
To show this, we carry out a motivation experiment with different learning rates for the predictor by setting the learning rate of the generator to be fixed at $\alpha_g = 0.0001$, and the learning rate of the predictor as $\alpha_p=\lambda\alpha_g$ ranging from $0.001$ ($\lambda=10$) to $0.0000067$ ($\lambda=1/15$). The Lipschitz constant is approximated by $\mathop{\max}\limits_{Z\in \mathcal{Z}_{train}}||\nabla_Z f_P(Z)||_2$ \cite{lip_calculate_by_grad_norm} (see Appendix~\ref{app:cal_lc}).
The results are shown in Figure~\ref{fig:lip with different pred lr}~(a). We can find that Lipschitz constant of the predictor exhibits explosive growth when $\alpha_p$ is higher than $\alpha_g$ (i.e., $\lambda>1$). And the Lipschitz constant are constrained to much smaller values when $\alpha_p$ is lower than $\alpha_g$ (i.e., $\lambda < 1$).

\subsection{The Proposed Method}
Although we conclude that small learning rate brings benefits for training RNP models, it is still hard to determine the optimal value. In fact, there exists a trade-off between the Lipschitz constant and the training speed where small learning rate will restrict the Lipschitz constant but will also slow down the training speed.  Inspired by the rationale sparsity constraint used in Equation~\ref{eqa:regular}, we in this paper propose a simple heuristic but empirically effective method of setting $\lambda$ for the learning rate of predictor such that the Lipschitz constant is reduced while still maintaining the training speed:
\begin{equation}\label{eqa:lambda}
    \lambda=\frac{||M||_1}{l},
\end{equation}
where $||M||_1$ is the number of selected tokens and $l$ is the length of the full text.
The generator can access the full input text, but the predictor can only get incomplete information contained in the rationale as a subset, which is only on average a proportion of $\frac{||M||_1}{l}$ of the full input text, so it seems reasonable to slow down the predictor by this ratio in order to make its decision more cautiously with incomplete information. We have also conducted experiments with $\lambda$ ranging form $5$ to $1/15$ in Section~\ref{sec:experiments}.

In Figure~\ref{fig:lip with different pred lr}~(b), we compare the approximate Lipschitz constants of the well trained predictors between RNP ($\lambda = 1$) and our model using asymmetric and adaptive learning rates ($\lambda=\frac{||M||_1}{l}$). We find that in all six aspects of the two datasets, our method gets much smaller Lipschitz constants compared to the vanilla RNP.  The results in Figure~\ref{fig:lip with different pred lr}~(b) are also consistent with the rationale quality results of $F1$ scores in Table~\ref{tab:beer} and Table~\ref{tab:hotel}, in which our method also gets higher $F1$ scores compared to the vanilla RNP in all six aspects. Again, this reflects that Lipschitz continuity does play an important role in the quality of the selected rationales. 

\begin{table}[t]
\caption{Statistics of datasets used in this paper}
   \centering
   \setlength\tabcolsep{2pt}
    \begin{tabular}{c l| c c| c c| c c c}
    \hline
         \multicolumn{2}{c|}{\multirow{2}{*}{Datasets}}&\multicolumn{2}{c|}{Train}&\multicolumn{2}{c|}{Dev}&\multicolumn{3}{c}{Annotation}  \\
         \multicolumn{2}{c|}{}& Pos&Neg&Pos&Neg&Pos&Neg&S\\
         \hline\hline
        \multirow{3}{*}{Beer}&Appearance&16891&16891 &6628&2103&923&13&18.5\\
        {}&Aroma&15169&15169&6579&2218&848&29&15.6\\
        {}&Palate&13652&13652&6740&2000&785&20&12.4\\
        \hline\hline
        \multirow{3}{*}{Hotel}&Location&7236&7236 &906&906&104&96&8.5\\
        {}&Service&50742&50742&6344&6344&101&99&11.5\\
        {}&Cleanliness&75049&75049&9382&9382&99&101&8.9\\
        \hline
    \end{tabular}
    
    \label{tab:dataset}
\end{table}

\section{Experiments}\label{sec:experiments}

\begin{table*}[t]
\setlength\tabcolsep{3.6pt}
\caption{Results on two standard benchmarks:  \emph{BeerAdvocate} and \emph{HotelReview} datasets. Each aspect is trained independently. $``*"$: results obtained from the paper of FR \cite{liufr}. $``$re-$"$: our reimplemented methods. The \underline{underlined} numbers are the second best results. }
    \centering
\subtable[BeerAdvocate]{   
\resizebox{1.99\columnwidth}{!}{
    \begin{tabular}{ c c |c c| c c |c|c c| c c |c |c c| c c |c }
\hline
\multicolumn{2}{c|}{\multirow{2}{*}{Methods}} & \multicolumn{5}{c|}{Appearance} & \multicolumn{5}{c|}{Aroma} & \multicolumn{5}{c}{Palate}\\
\cline{3-17}
\multicolumn{2}{c|}{} &S& Acc & P & R &\multicolumn{1}{|c|}{F1} &S& Acc & P & R &\multicolumn{1}{|c|}{F1} &S& Acc& P & R &\multicolumn{1}{|c}{F1}\\
\hline
\multicolumn{2}{c|}{DMR*} &18.2&-& 71.1 &70.2 & 70.7 &15.4&-&59.8 & 58.9 & 59.3 &11.9&-&53.2 &50.9 & 52.0\\
\multicolumn{2}{c|}{A2R*} & 18.4 & 83.9 & 72.7 &72.3 & 72.5 & 15.4 & 86.3 & 63.6 & 62.9&63.2&12.4&81.2&57.4&57.3&57.4\\
\multicolumn{2}{c|}{FR*} &18.4&87.2&{82.9}&{82.6}&\underline{82.8}&15.0&88.6&{74.7}&{72.1}&\underline{73.4}&12.1&89.7&\textbf{67.8}&66.2&\underline{67.0}\\
\multicolumn{2}{c|}{re-RNP} &18.2& 83.3 & 73.8 & 72.7 & 73.2 &16.0&85.2&64.1 &65.9& 64.9 & 13.0 & 85.2 & 60.1 & 63.1 & 61.5\\
\multicolumn{2}{c|}{DR(ours)} &18.6&85.3&\textbf{84.3}&\textbf{84.8}&\textbf{84.5}&15.6&87.2&\textbf{77.2}&\textbf{77.5}&\textbf{77.3}&13.3&85.7&{65.1}&\textbf{69.8}&\textbf{67.4}
  \\\hline
\end{tabular}
}
    
    \label{tab:beer}
}

\subtable[HotelReview]{
    \resizebox{1.99\columnwidth}{!}{
    \begin{tabular}{c c c |c c| c c |c|c c| c c |c |c c| c c |c }
\hline
\multicolumn{3}{c|}{\multirow{2}{*}{Methods}} & \multicolumn{5}{c|}{Location} & \multicolumn{5}{c|}{Service} & \multicolumn{5}{c}{Cleanliness}\\
\cline{4-18}
\multicolumn{3}{c|}{} &S& Acc & P & R &\multicolumn{1}{|c|}{F1} &S& Acc & P & R &\multicolumn{1}{|c|}{F1} &S& Acc& P & R &\multicolumn{1}{|c}{F1}\\
\hline

\multicolumn{3}{c|}{DMR*} & 10.7 & - & 47.5 &{60.1} & 53.1 & 11.6 & - & 43.0 &43.6&43.3&10.3&-&31.4&36.4&33.7\\
\multicolumn{3}{c|}{A2R*} & 8.5 &87.5 &43.1 &43.2 & 43.1 &11.4 & 96.5 & 37.3 & 37.2&37.2&8.9&94.5&33.2&33.3&33.3\\
\multicolumn{3}{c|}{FR*} &9.0&93.5 & \textbf{55.5}&58.9&\textbf{57.1}&11.5& 94.5&{44.8}&{44.7}&\underline{44.8}&11.0&96.0&{34.9}&{43.4}&\underline{38.7}\\
\multicolumn{3}{c|}{RNP*} & 8.8& 97.5 & 46.2 &48.2 &47.1 &11.0 &97.5 & 34.2 & 32.9&33.5&10.5&96.0&29.1&34.6&31.6\\
\multicolumn{3}{c|}{DR(ours)} &9.6&96.5&{53.6}&\textbf{60.9}&\underline{57.0}&11.5&{96.0}&\textbf{47.1}&\textbf{47.4}&\textbf{47.2}&10.0&{97.0}&\textbf{39.3}&\textbf{44.3}&\textbf{41.8}
  \\\hline
\end{tabular}
}
  
    \label{tab:hotel}
}
\end{table*}

\subsection{Datasets}
Following \citet{dmr} and \citet{liufr}, we consider two widely used datasets for rationalization tasks. \textbf{BeerAdvocate} \citep{beer} is a multi-aspect sentiment prediction dataset on reviewing beers. 
Following previous works \citep{car,dmr,interlocking,liufr}, we use the subsets decorrelated by \citet{emnlp/LeiBJ16} and binarize the labels as \citet{2018rationalegumble} did.
\textbf{HotelReview} \citep{hotel} is another multi-aspect sentiment prediction dataset on reviewing hotels. The dataset contains reviews of hotels from three aspects including location, cleanliness, and service. Each review has a rating on a scale of 0-5 stars. We binarize the labels as \cite{2018rationalegumble} did. Both datasets contain human-annotated rationales on the annotation (test) set only. The statistics of the datasets are in Table~\ref{tab:dataset}. $Pos$ and $Neg$ denote the number of positive and negative examples in each set. $S$ denotes the average percentage of tokens in human-annotated rationales to the whole texts.

We preprocess both datasets in the same way as FR \cite{liufr} for a fair comparison and the details are in Appendix~\ref{app:datasets}.

\subsection{Baselines and Implementation Details }
The main baseline for direct comparison is the original cooperative rationalization framework RNP \citep{emnlp/LeiBJ16}, as RNP and our DR match in selection granularity, optimization algorithm and model architecture, which helps us to focus on our claims rather than some potential unknown mechanisms. 
To show the competitiveness of our method, we also include several recently published models that achieve state-of-the-art results: DMR \cite{dmr}, A2R \cite{interlocking} and FR \cite{liufr}, all of which have been discussed in detail in Section~\ref{sec:related}. 

Experiments in recent works show that it is still a challenging task to finetune large pretrained language models on the RNP cooperative framework \cite{danqi,liufr}. For example, Table~\ref{tab:models with bert} shows that several improved rationalization methods fail to  find the true rationales when using pretrained models (e.g., ELECTRA-small \cite{electra} and BERT-base \cite{bert}) as the encoders of the players. To make a fair comparison, we take the same setting as previous works do \cite{dmr,interlocking,liufr}. We use one-layer 200-dimension bi-directional gated recurrent units (GRUs) \citep{gru} followed by one linear layer for each of the players, and the word embedding is 100-dimension Glove \citep{glove}. The optimizer is Adam \citep{adam}. The  reparameterization trick for binarized sampling is Gumbel-softmax \citep{2016gumble,2018rationalegumble}, which is also the same as FR. We take the results when the development set gets the highest prediction accuracy. 
To show the competitiveness of our DR, we further conduct some experiments with pretrained language models as a supplement to the main experiments.

\textbf{Metrics}. Following previous works \cite{dmr,interlocking,liufr}, we focus on the rationale quality, which is measured by the overlap between the model-selected tokens and human-annotated tokens. $P,R,F1$ indicate the precision, recall, and F1 score, respectively. $S$ indicates the average 
percentage (sparsity) of selected tokens to the whole texts. $Acc$ indicates the predictive accuracy.

\subsection{Results}\label{sec:result}

\textbf{Comparison with state-of-the-arts}.
Table~\ref{tab:beer} and ~\ref{tab:hotel} show the main results compared to previous methods on two standard benchmarks. 
In terms of $F1$ score, our approach outperforms the best available methods in five out of six aspects of the two datasets.
In particular, as compared to our direct baseline RNP, we get improvements of more than $10\%$ in terms of F1 score in four aspects (i.e., Appearance, Aroma, Service, Cleanliness).
The improvement shows that the heuristic of setting $\lambda$ in our DR is a very strong one. Figure~\ref{fig:lip with different pred lr}~(b) shows the Lipschitz constants of well-trained RNP and DR on different datasets.
Table~\ref{tab:examples} shows some visualized examples of rationales from RNP and our DR. We also provide some failure cases of our DR in Appendix~\ref{app:fail} to pave the way for future work.

\begin{table*}[t]
\setlength\tabcolsep{3.6pt}
\caption{Results of methods with low sparsity on \emph{BeerAdvocate}. $``*"$ and $``**"$: results obtained from the papers of CAR \cite{car}  and DMR \cite{dmr}, respectively.  }

        \resizebox{1.99\columnwidth}{!}{
\begin{tabular}{c c c |c c| c c |c|c c| c c |c |c c| c c |c }
\hline
\multicolumn{3}{c|}{\multirow{2}{*}{Methods}} & \multicolumn{5}{c|}{Appearance} & \multicolumn{5}{c|}{Aroma} & \multicolumn{5}{c}{Plate}\\
\cline{4-18}
\multicolumn{3}{c|}{} &S& Acc & P & R &\multicolumn{1}{c|}{F1} &S& Acc & P & R &\multicolumn{1}{c|}{F1} &S& Acc& P & R &\multicolumn{1}{c}{F1}\\
\hline

\multicolumn{3}{c|}{RNP$^{*}$} & 11.9&-& 72.0 & 46.1 &56.2 &10.7 &- & 70.5 & 48.3&57.3&10.0&-&53.1&42.8&47.5\\
\multicolumn{3}{c|}{CAR$^{*}$} & 11.9 & - & 76.2 & 49.3 & 59.9 & 10.3 & - & 50.3 &33.3&40.1&10.2&-&56.6&46.2&50.9\\
\multicolumn{3}{c|}{DMR$^{**}$} & 11.7 &- &83.6 &52.8 & 64.7 &11.7 &-& 63.1 & 47.6 & 54.3&10.7&-&55.8&48.1&51.7\\
\multicolumn{3}{c|}{DR(ours)} &11.9&81.4&\textbf{86.8} & \textbf{55.9}&\textbf{68.0}&11.2&80.5& \textbf{70.8}&\textbf{57.1}&\textbf{63.2}&10.5&81.4&\textbf{71.22}&\textbf{60.2}&\textbf{65.3}
  \\\hline
\end{tabular}
}

    \label{tab:beer_lowsp}

\end{table*}

\begin{figure}[t]
\centering
    \includegraphics[width=0.996\columnwidth]{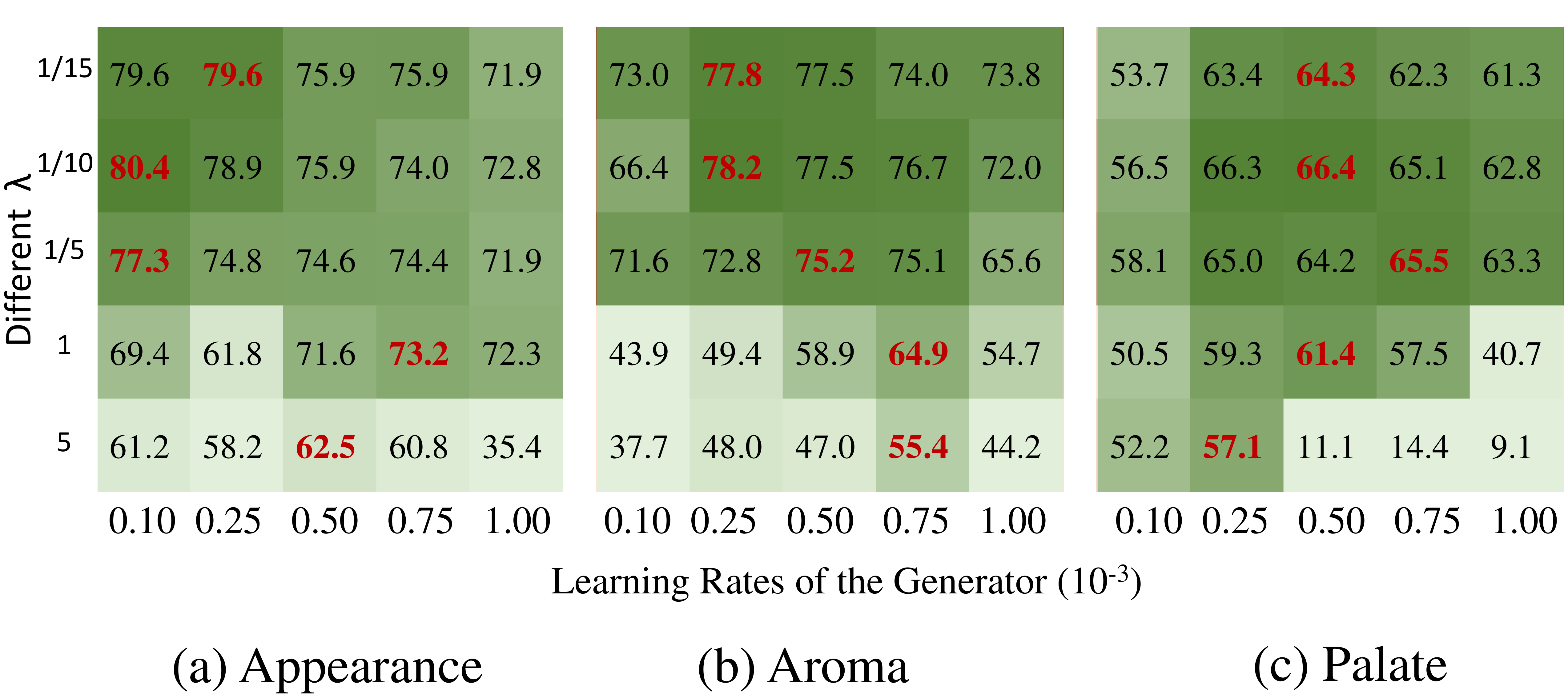}
  \caption{Results with different values of $\lambda$ in the three aspects of  \emph{BeerAdvocate}. The values in these cells are F1 scores.}
  \label{fig:ex_lambda}

\end{figure}

\textbf{Results with different $\lambda$ values}. Since we do not have a rigorous theoretical analysis for choosing the best $\lambda$, in Figure~\ref{fig:ex_lambda} we run the experiments with a wide range of values of $\lambda$ to verify our claim that lowering the learning rate of predictor can alleviate degeneration. We take five different values of $\lambda$: 1/15, 1/10, 1/5, 1, 5 (shown along the vertical axis) and five different learning rates for the generator ranging from $0.0001$ to $0.001$ (with corresponding ratios shown along the horizontal axis). The values in the cells are $F1$ scores reflecting rationale quality. 
The experimental results demonstrate that, given a vanilla RNP model with the setting of $\lambda=1$, making the learning rate of its predictor lower than that of its generator, i.e., lowering $\lambda$ to some value smaller than 1, can always get rationales much better than the original RNP. 
And it seems that the results are not too sensitive to the value of $\lambda$ as long as the value of $\lambda$ is smaller than 1. 
We also compare the results of our method DR ($\lambda=\frac{||M||_1}{l}$) shown in Table~\ref{tab:beer} with the corresponding best results in Figure~\ref{fig:ex_lambda}, and find that our heuristic setting of $\lambda$ in Equation~\ref{eqa:lambda} is a reasonably good option that is empirically close to the optimal solution.

\textbf{Comparison with baselines in low sparsity}.
To show the robustness of our method, we also conduct an experiment where the sparsity of selected rationales is very low, which is the same as the sparsity used in CAR \citep{car} and DMR \cite{dmr}. The results are shown in Table~\ref{tab:beer_lowsp}. 
We still significantly outperform the baselines.

\textbf{Inducing degeneration with skewed predictor.} To show that even if the predictor overfits to trivial patterns, our DR can still escape from degeneration, we conduct the same synthetic experiment that gives the predictor very poor initialization and deliberately induces degeneration as \citet{interlocking} did. We first pretrain the predictor separately using only the first sentence of input text, and then cooperatively train the predictor initialized with the pretrained parameters and the generator randomly initialized using normal input text. “\emph{skew}$k$” means that the predictor is pre-trained for $k$ epochs. For a fair comparison, we keep the pre-training process the same as in A2R: we use a batch-size of 500 and a learning rate of 0.001. 
More details of this experiment can be found in Appendix~\ref{app:skewpred}.

The results are shown in Table~\ref{tab:predskew}. The results of RNP and A2R are obtained from A2R \citep{interlocking}, and results of FR are also copied from its original paper \cite{liufr}. For all the settings, we outperform all the three methods.
Especially, for the relatively easy task on \emph{Aroma}, the performance of DR is hardly affected by this poor initialization. And for the relatively hard task on \emph{Palate}, it is only slightly affected, while RNP and A2R can hardly work and FR is also significantly affected. It can be concluded that DR is much more robust than the other two methods in this situation.

\textbf{Inducing degeneration with skewed generator}. It is a synthetic experiment with a special initialization that induces degeneration from the perspective of skewed generator, which is designed by \citet{liufr}. 
We pretrain the generator separately using the text classification label as the mask label of the first token. In other words, for texts of class 1, we force the generator to select the first token, and for texts of class 0, we force the generator not to select the first token. So, the generator learns the category implicitly by whether the first token is chosen and the predictor only needs to learn this position information to make a correct prediction.

All the hyperparameters are the same as those of the best results of the corresponding models in Table~\ref{tab:beer}. We use \emph{Beer-Palate} because A2R and FR show that Palate is harder than other aspects. $k$ in “skew$k$” denotes the threshold of the skew: we pretrain the generator as a special classifier of the first token for a few epochs until its prediction accuracy is higher than $k$. Since the accuracy increases rapidly in the first a few epochs, obtaining a model that precisely achieves the pre-defined accuracy is almost impossible. So, we use “$Pre\_acc$” to denote the actual prediction accuracy of the generator-classifier when the pre-training process stops. Higher “$Pre\_acc$” means easier to degenerate.

The results are shown in Table~\ref{tab:skew generator}. In this case, RNP fails to find the human-annotated rationales and FR is also significantly affected when the skew degree is high. While our DR is much less affected, which shows the robustness of our DR under this case.

\begin{table*}[t]
\setlength\tabcolsep{3.6pt}
 \caption{Results of skewed predictor that induces degeneration on \emph{BeerAdvocate}. $``*"$ and $``**"$: results obtained from the papers of A2R \cite{interlocking}  and FR \cite{liufr}, respectively.} 
    \centering
    \resizebox{1.99\columnwidth}{!}{
    \begin{tabular}{c |c |c| c c |c|c| c c |c|c| c c |c|c| c c |c}
    \hline
    \multirow{2}{*}{Aspect} &\multirow{2}{*}{Setting}& \multicolumn{4}{c|}{RNP*} & \multicolumn{4}{c|}{A2R*}& \multicolumn{4}{c|}{FR**}&\multicolumn{4}{c}{DR(ours)} \\
\cline{3-18}
{}&{} &Acc&P & R &F1&Acc & P & R &F1&Acc&P & R &F1&Acc & P & R &F1\\
\hline
\multicolumn{1}{c|}{{\multirow{3}{*}{Aroma}}} &\multicolumn{1}{c|}{skew10} &82.6&68.5 &63.7 & 61.5 &84.5&\textbf{78.3}&70.6&69.2& 87.1 &73.9&{71.7}&{72.8}& 85.0 &77.3&\textbf{75.7}&\textbf{76.5}\\
\multicolumn{1}{c|}{}&\multicolumn{1}{c|}{skew15} &80.4&54.5& 51.6&49.3 &81.8&58.1&53.3&51.7& 86.7 &{71.3}&{68.0}&{69.6}& 85.4 &\textbf{76.1}&\textbf{77.2}&\textbf{76.6}\\
\multicolumn{1}{c|}{}&\multicolumn{1}{c|}{skew20} &76.8&10.8 & 14.1 &11.0 &80.0&51.7&47.9&46.3&85.5 &{72.3}&{69.0}&{70.6}&85.5 &\textbf{77.3}&\textbf{76.2}&\textbf{76.8}\\
\hline
\multicolumn{1}{c|}{{\multirow{3}{*}{Palate}}} &\multicolumn{1}{c|}{skew10} &77.3&5.6 &7.4 & 5.5 &82.8&50.3&48.0&45.5&75.8&{54.6}&{61.2}&{57.7} &85.8&\textbf{67.7}&\textbf{68.6}&\textbf{68.2}\\
\multicolumn{1}{c|}{}&\multicolumn{1}{c|}{skew15} &77.1&1.2 & 2.5 & 1.3 &80.9&30.2&29.9&27.7&81.7&{51.0}&{58.4}&{54.5}&83.9&\textbf{66.3}&\textbf{66.7}&\textbf{66.5}\\
\multicolumn{1}{c|}{}&\multicolumn{1}{c|}{skew20} &75.6&0.4 & 1.4 & 0.6 &76.7&0.4&1.6&0.6& 83.1 &{48.0}&{58.9}&{52.9}& 85.0 &\textbf{59.4}&\textbf{62.6}&\textbf{61.0}\\
\hline
    \end{tabular}
    }
 
     \label{tab:predskew}

\end{table*}

\begin{table*}[t]
\setlength\tabcolsep{3pt}
\caption{Results of skewed generator that induces degeneration in the \emph{Palate} aspect of \emph{BeerAdvocate}. $``*"$: results obtained from the paper of FR.}
    \centering
    \resizebox{1.99\columnwidth}{!}{
    \begin{tabular}{c |c c c| c c| c| c c c| c c| c| c c c| c c| c}
    \hline
    \multicolumn{1}{c|}{\multirow{2}{*}{Setting}} & \multicolumn{6}{c|}{RNP*} & \multicolumn{6}{c|}{FR*}& \multicolumn{6}{c}{DR(ours)}\\
\cline{2-19}
\multicolumn{1}{c|}{} &Pre\_acc&S&Acc & R & R &F1&Pre\_acc&S&Acc & P & R &F1&Pre\_acc&S&Acc & P & R &F1\\
\hline
\multicolumn{1}{c|}{skew65.0} &66.6&14.0&83.9 &40.3 &45.4&42.7&66.3&14.2&81.5&{59.5}&\textbf{67.9}&\textbf{63.4}&68.5&12.5&81.9&\textbf{60.9}&{61.1}&{61.0}\\
\multicolumn{1}{c|}{skew70.0} &71.3&14.7&84.1&10.0 & 11.7 & 10.8&70.8&14.1&88.3&{54.7}& {62.1} &{58.1}&70.3&12.8&84.2&\textbf{61.8}& \textbf{63.6} &\textbf{62.6}\\
\multicolumn{1}{c|}{skew75.0} &75.5 & 14.7 &87.6 &8.1&9.6&8.8&75.6&13.1&84.8&{49.7}& {52.2} &{51.0}&75.5&12.8&81.8&\textbf{58.8}& \textbf{60.6} &\textbf{59.7}\\
\hline
    \end{tabular}
    }
    
    \label{tab:skew generator}

\end{table*}

\textbf{Experiments with pretrained language models.}  In the field of rationalization, researchers generally focus on frameworks of the models and the methodology rather than engineering SOTA. The methods most related to our work do not use BERT or other pre-trained encoders \cite{car,invarant,dmr,rethinking,interlocking,liufr}. Experiments in some recent work \cite{danqi,liufr} indicate that there are some unknown obstacles making it hard to finetune large pretrained models on the self-explaining rationalization framework. For example, Table~\ref{tab:models with bert} shows that two improved rationalization methods (VIB \cite{informationbottle} and SPECTRA \cite{spectral}) and the latest published FR all fail to find the informative rationales when replacing GRUs with pretrained language models. To eliminate potential factors that could lead to an unfair comparison, we adopt the most widely used GRUs as the encoders in our main experiments, which can help us focus more on our claims themselves rather than unknown tricks.

But to show the competitiveness of our DR, we also provide some experiments with pretrained language models as the supplement. Due to limited GPU resources, we adopt the relatively small ELECTRA-small \cite{electra} in all three aspects of \emph{BeerAdvocate} and the relatively large BERT-base \cite{bert} in the \emph{Appearance} aspect. We compare our DR with the latest SOTA FR \cite{liufr}.
The results with BERT-base are shown in Table~\ref{tab:models with bert} and the results with ELECTRA-small are shown in Table~\ref{tab:DR AND FR with electra}.  Although DR sometimes still performs worse than the models with GRUs, it makes a great progress in this line of research as compared to the previous methods like VIB \cite{informationbottle}, SPECTRA \cite{spectral} and FR \cite{liufr}. 
Although we still do not get as good rationales as using GRUs, we are making great strides compared to previous methods.

\begin{table}[t]
 \caption{The rationale quality (F1 score) of models trained with different encoders. $``$*$"$: The results of VIB \cite{informationbottle} and SPECTRA \cite{spectral} are from \citet{danqi}. $``$**$"$: The results of RNP and FR \cite{liufr} are from \citet{liufr}. The dataset is \emph{Beer-Appearance}.}
    \centering
    \resizebox{0.8\columnwidth}{!}{
    \begin{tabular}{c|c|c|c}
    \hline
         Method& GRU & ELECTRA&BERT  \\
         \hline
         VIB*& -&-&20.5 \\
         \hline
         SPECTRA* & -&-&28.6\\
         \hline
         RNP**&72.3&13.7&14.7\\
         \hline
         FR**&82.8&14.6&29.8\\
         \hline
         DR(ours) &84.5&87.2&85.0\\
         \hline
    \end{tabular}
   }
    \label{tab:models with bert}
\end{table}

\begin{table*}[t]
\setlength\tabcolsep{3.6pt}
\caption{Results of methods using pretrained ELECTRA as the encoder. The dataset is \emph{BeerAdvocate}.  }

        \resizebox{1.99\columnwidth}{!}{
\begin{tabular}{c c c |c c| c c |c|c c| c c |c |c c| c c |c }
\hline
\multicolumn{3}{c|}{\multirow{2}{*}{Methods}} & \multicolumn{5}{c|}{Appearance} & \multicolumn{5}{c|}{Aroma} & \multicolumn{5}{c}{Plate}\\
\cline{4-18}
\multicolumn{3}{c|}{} &S& Acc & P & R &\multicolumn{1}{c|}{F1} &S& Acc & P & R &\multicolumn{1}{c|}{F1} &S& Acc& P & R &\multicolumn{1}{c}{F1}\\
\hline

\multicolumn{3}{c|}{FR} & 16.3&86.5& 19.1 & 17.0 &18.0&14.8 &85.9& 58.6 & 54.8&56.7&11.2&78.0&12.0&10.7&11.3\\
\multicolumn{3}{c|}{DR(ours)} &17.4&93.0&\textbf{89.4} & \textbf{85.0}&\textbf{87.2}&14.5&89.5& \textbf{79.2}&\textbf{72.4}&\textbf{75.7}&12.5&88.9&\textbf{61.1}&\textbf{60.5}&\textbf{60.8}
  \\\hline
\end{tabular}
}

    \label{tab:DR AND FR with electra}

\end{table*}

\begin{table*}[t]
\caption{Examples of generated rationales. Human-annotated rationales are \underline{underlined}. Rationales from RNP and DR are highlighted in \textcolor{red}{red} and \textcolor{blue}{blue} respectively. }
\footnotesize
\begin{tabularx}{\linewidth}{X X}
	\hline\hline
	\makecell[c]{RNP} &\makecell[c]{DR(ours)}\\\hline\hline
	\textbf{Aspect:} Beer-Aroma &\textbf{Aspect:} Beer-Aroma     \\\textbf{Label:} Positive, \textbf{Pred:} Positive&\textbf{Label:} Positive, \textbf{Pred:} Positive\\
 \textbf{Text:} this beer poured a hypnotically deep amber color almost as though it wanted to be an american brown . near white head that faded quickly way darker than i was expecting . \  \underline{hops \textcolor{red}{nose} but not nearly as aggressive as most ipas . so far not what i think} \underline{of an ipa .} the taste is malty at the very beginning then followed by a punch of hops that hits you right in the back of the jaw . exactly what it should be . the finish is especially bitter and hoppy . the mouthfeel is good ; moderate carbonation and smooth . poignant and bitter in a notable \textcolor{red}{way} . a strange ipa but an outstanding one . 
		&  \textbf{Text:} this beer poured a hypnotically deep amber color almost as though it wanted  to be an american brown . near white head that faded quickly way darker than i was  \textcolor{blue}{expecting} \textcolor{blue}{.} \  \underline{\textcolor{blue}{hops} \textcolor{blue}{nose} \textcolor{blue}{but} \textcolor{blue}{not} \textcolor{blue}{nearly} \textcolor{blue}{as} \textcolor{blue}{aggressive} \textcolor{blue}{as} \textcolor{blue}{most} \textcolor{blue}{ipas} \textcolor{blue}{.} \textcolor{blue}{so} \textcolor{blue}{far} \textcolor{blue}{not} \textcolor{blue}{what} \textcolor{blue}{i} \textcolor{blue}{think}} \underline{\textcolor{blue}{of} \textcolor{blue}{an} ipa .} the taste is malty at the very beginning then followed by a punch of hops that hits you right in the back of the jaw . exactly what it should be . the finish is especially bitter and hoppy . the mouthfeel is good ; moderate carbonation and smooth . poignant and bitter in a notable way . a strange ipa but an outstanding one . \\
    \hline
    \hline
    \textbf{Aspect:} Beer-Palate &\textbf{Aspect:} Beer-Palate    \\\textbf{Label:} Positive, \textbf{Pred:} Positive&\textbf{Label:} Positive, \textbf{Pred:} Positive\\
 \textbf{Text:} on tap at pope in philly pours a completely clouded golden color with some yellow hues . thick , frothy white head retains well on top , fading into some layers of film with some rings of lacing on the glass spicy and citrusy on the nose . some lemon peel and cracked pepper with a hint of orange sweetness . barely any alcohol with some soft brett funk and yeast splash of floral hops up front with peppery spice . some herbal notes followed by citrus juiciness and lemon peel . a little sweet in the middle from the fruit . hints of [unknown] funk and belgian yeast , but not quite tart . finishes \textcolor{red}{a} little \textcolor{red}{spicy} \  \underline{medium \textcolor{red}{body} , higher carbonation , refreshing and tingly on the palate .}  a solid saison , although i was hoping for more brett presence . 
		&  \textbf{Text:} on tap at pope in philly pours a completely clouded golden color with some yellow hues . thick , frothy white head retains well on top , fading into some layers of film with some rings of lacing on the glass spicy and citrusy on the nose . some lemon peel and cracked pepper with a hint of orange sweetness . barely any alcohol with some soft brett funk and yeast splash of floral hops up front with peppery spice . some herbal notes followed by citrus juiciness and lemon peel . a little sweet in the middle from the fruit . hints of [unknown] funk and belgian yeast , but not quite tart . \textcolor{blue}{finishes a little spicy} \ \underline{\textcolor{blue}{medium} \textcolor{blue}{body} \textcolor{blue}{,} \textcolor{blue}{higher} \textcolor{blue}{carbonation} \textcolor{blue}{,} \textcolor{blue}{refreshing} \textcolor{blue}{and} \textcolor{blue}{tingly} \textcolor{blue}{on} \textcolor{blue}{the} \textcolor{blue}{palate} \textcolor{blue}{.}}  a solid saison , although i was hoping for more brett presence . \\
 
		\hline\hline

\end{tabularx}
\label{tab:examples}
\end{table*}

\begin{figure*}[t]
    \centering
    \subfigure[]{
    \includegraphics[width=0.49\columnwidth]{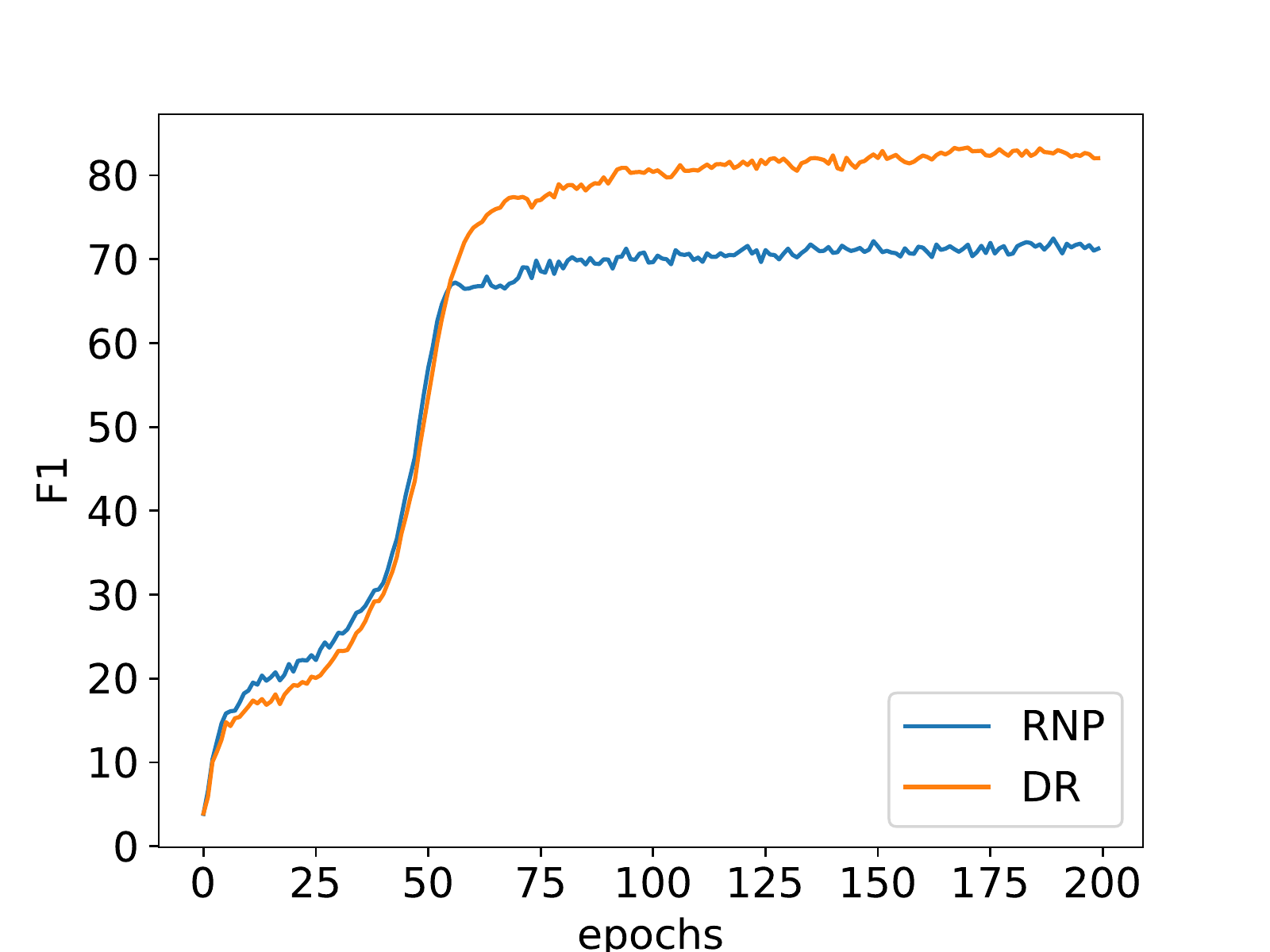}
        }
      \subfigure[]{
        \includegraphics[width=0.49\columnwidth]{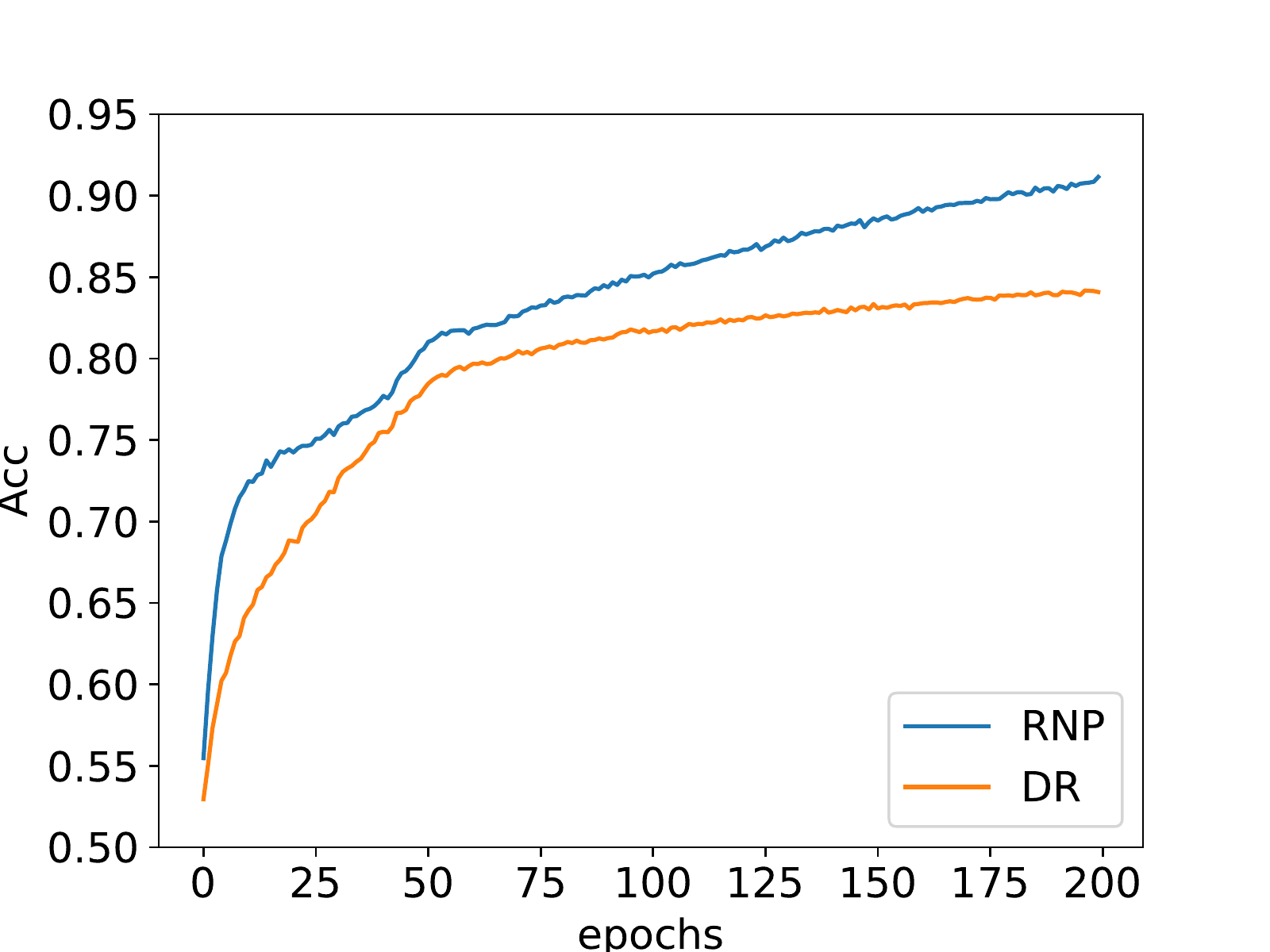}
        }
        \subfigure[]{
    \includegraphics[width=0.49\columnwidth]{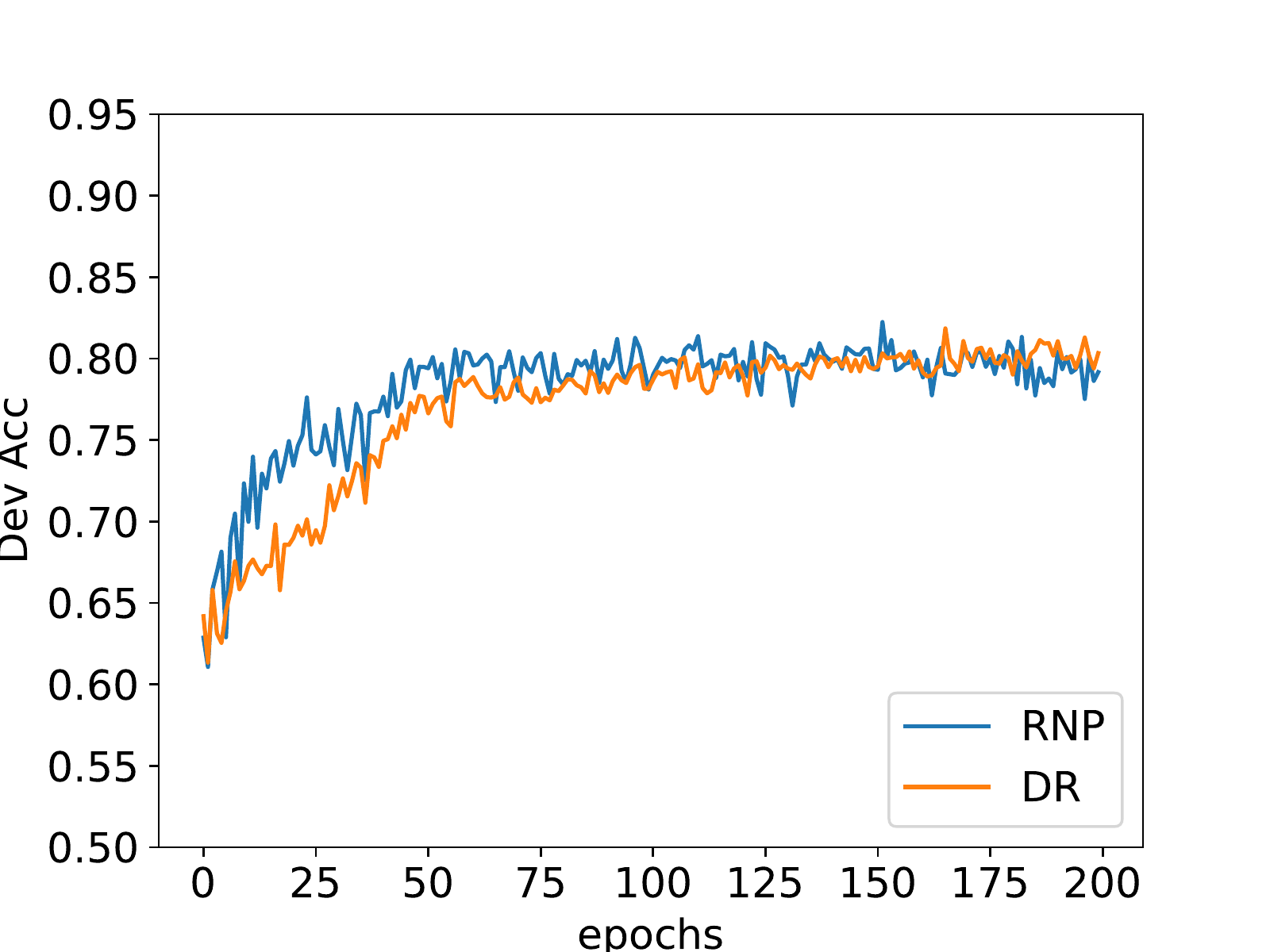}
        }
    \subfigure[]{
        \includegraphics[width=0.49\columnwidth]{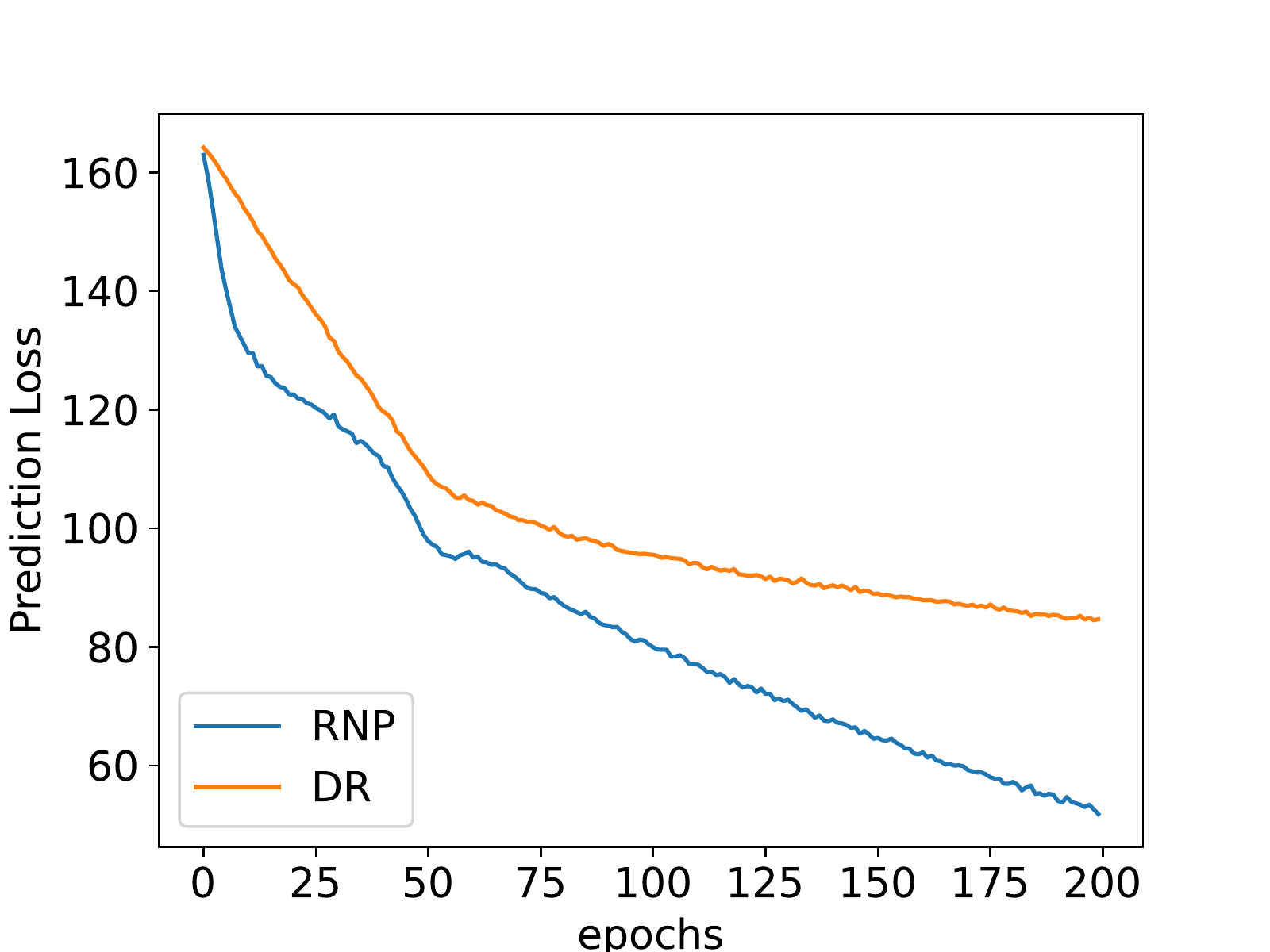}
    }

  \caption{The training process of RNP and our DR on \emph{Beer-Appearance}.}
  \label{fig:time}
 
\end{figure*}

\textbf{Time efficiency and overfitting}.
At first glance, it seems that lowering the learning rate of the predictor might slow down the convergence of the training process.
However, the experiments show that convergence of our model is not slowed down at all. Figure~\ref{fig:time}~(a) shows the evolution of the $F1$ scores over the training epochs. All the settings are the same except that we use $\lambda=\frac{||M||_1}{l}$ while RNP uses $\lambda=1$. 
Figure~\ref{fig:time}~(b) shows that the training accuracy of RNP grows fast in the beginning when limited rationales have been sampled by the generator, and reflects that the predictor is trying to overfit to these randomly sampled rationales. Figure~\ref{fig:time}~(c) shows that, although RNP gets a very high accuracy in the training dataset, it does not get accuracy higher than our method in the development dateset, and also indicates the fact of overfitting. The prediction loss in Figure~\ref{fig:time}~(d) reflects a similar phenomenon.

We show more results about the time efficiency and overfitting phenomenon on \emph{Beer-Aroma} and \emph{Beer-Palate} in Figure~\ref{fig:time_more} of the Appendix. We see that we have better or close time efficiency compared to Vanilla RNP and we always get much less overfitting.

\section{Conclusion and future work }
In this paper, we first bridge the degeneration in cooperative rationalization with the predictor's Lipschitz continuity. Then we analyze the two players' coordination so as to propose a simple but effective method to guide the predictor to converge with good Lipschitz continuity. Compared to existing methods, we do not impose complex regularizations or rigid restrictions on the model parameters of the RNP framework, so that the predictive capacity is not affected. Although there is no rigorous theoretical analysis on how to choose the best $\lambda$, the method to obtain good Lipschitz continuity in this paper is simple but empirically effective, which paves the way for future research. Other methods that can help to obtain networks with good Lipschitz continuity can be also explored to promote rationalization research in the future.
 
Given the versatility of the self-explaining rationalization framework, our proposed methods show significant potential for application across diverse fields such as multi-aspect recommender systems \citep {deng2023multi} and computer vision \citep{GDM}. 

\section{Limitations}

One limitation may be that how to apply this simple learning rate strategy to other variants of this type of two-player rationalization requires further exploration, which we leave as future work.

Another limitation is that the obstacles in utilizing powerful pretrained language models under the rationalization framework remain mysterious. Though we have made some progress in this direction, the empirical results with pretrained models don't show significant improvements as compared to those with GRUs. Further efforts should be made in the future. However, it's somewhat beyond the scope of this paper, and we leave it as the future work.

\section*{Acknowledgements}
This work is supported by National Natural Science Foundation of China under grants U1836204, U1936108, 62206102,  and Science and Technology Support Program of Hubei Province under grant 2022BAA046. This paper is a collaborative work between the Intelligent and Distributed Computing Laboratory at Huazhong University of Science and Technology, and \href{https://www.iwudao.tech/}{iWudao Tech}. We thank the anonymous reviewers for their valuable comments on improving the quality of this paper.


\bibliographystyle{ACM-Reference-Format}
\balance
\bibliography{sample-base}

\clearpage

\appendix

\section{Experimental Setup}
\subsection{Datasets}\label{app:datasets}
\textbf{BeerAdvacate} Following \citet{car,dmr,interlocking}, we consider a classification setting by treating reviews with ratings $\leq$ 0.4 as
negative and $\geq$ 0.6 as positive. Then we randomly select examples from the original training set to
construct a balanced set.

\textbf{Hotel Reviews}
Similar to BeerAdvacate, we treat reviews with ratings $<$ 3 as
negative and $>$ 3 as positive.

\subsection{Implementation details}\label{app: implementation details}
We reimplement RNP with updated short and coherent regularizers (Equation~\ref{eqa:regular}) and reparameterization trick (gumbel-softmax) to make RNP and our DR match in all the details except that DR uses asymmetric learning rates.
The early stop is conducted according to the predictive accuracy on the development set.

\subsection{The setup of Figure~\ref{fig:distance}}\label{app:distance}
We first give the formal definition of the properties that a distance metric should have:
\begin{definition}\label{def:properties of distance}
    We call $d$ is a distance metric if and only if it satisfies the four basic properties:
\begin{equation}
    \begin{aligned}
        &d(A,B)\geq 0, \\
        &d(A,B)=0 \ \text{iff} \  A=B,\\
        &d(A,B)=d(B,A),\\
        &d(A,B)\leq d(A,C)+d(B,C).
    \end{aligned}
\end{equation}
\end{definition}
We denote the set of all the possible distance metric as $\mathcal{M}$.

The specific distance in Figure~\ref{fig:distance} is defined as: 
\begin{definition}
\label{def:average distance}
For two text sequences $A=a_{1:n}$, $B=b_{1:m}$, where $a_i,b_i\in R^d$ are the word embeddings (Glove-100 in this paper). We define a distance between them as
\begin{equation}
    d_t(A,B)=||\frac{1}{n}\sum\limits_{i=1}^{n}a_i-\frac{1}{m}\sum\limits_{i=1}^{m}b_i||_2,
\end{equation}
\end{definition}
where we use the average because different sentences have different lengths. It is easy to prove that this definition satisfies the four basic properties implied in Definition~\ref{def:properties of distance}. It's worth noting that our theoretical derivations doesn't involve any specific distance definition, instead, they hold for any $d\in\mathcal{M}$. Definition~\ref{def:average distance} is only used in our quantitative experiments.

We uses the Hotel dataset because the Beer dataset is not balanced (see Table~\ref{tab:dataset}). We first randomly sample the same number of words as the human-annotated rationale (GR in Figure~\ref{fig:distance}) from each text in the annotation set as the uninformative rationale (RR in Figure~\ref{fig:distance}). The word vector is 100-dimension GloVe. 
Note that the original $Z$ is a matrix with shape $\mathcal{R}^{b\times d}$, where $b$ is the number of tokens in $Z$ and $d$ is the dimension of the word vector. $b$ varies according to different $Z$. To compute the norm, we need to unify rationales of different lengths into a vector of the same length.
 We first obtain the representation of a rationale by averaging the word vectors of all its words.
 Formally, for a rationale $Z=[z_1,z_2,\cdots,z_b]$ where $z_i\in \mathcal{R}^d$ is the $i$-th token, we represent $Z$ as a $d$-dimensional vertor by $Z=\frac{z_1+z_2+\cdots+z_b}{b}$.  
 Now we get $Z\in \mathcal{R}^{d}$. Then, we obtain the centroids of the categories by averaging over the rationales in the same category. Finally, we obtain the average distance between the two categories by measuring distance between corresponding centroids with the norm of different orders.

\subsection{Calculating the approximate Lipschitz constant through sampling gradient norm}\label{app:cal_lc}
As implied in \cite{lip_calculate_by_grad_norm,iclr18evaluate}, the Lipschitz constant can be calculated by the gradient norm.
\begin{lemma}
If a function $f(\cdot): \mathcal{R}^n\xrightarrow{}\mathcal{R}^1$ is Lipschitz continuous on $\mathcal{X}\subset \mathcal{R}^n$, then, the Lipschitz constant corresponding to Equation~\ref{eqa:lip_c} can be calculated through
\begin{equation}
    L_c = \mathop{\max}\limits_{X \in \mathcal{X}}{||\nabla f(X)||_q }, \ s.t.\  \frac{1}{p}+\frac{1}{q}=1
\end{equation}
where $\nabla f(X)=(\frac{\partial f(X)}{\partial x_1},\cdots,\frac{\partial f(X)}{\partial x_n})^\top$.
\end{lemma}
A usual setting is that $p=q=2$.
Here we calculate the Lipschitz constant of the predictor whose input is a rationale and the output is a value ranging in $(0,1)$.
It is intractable to calculate the maximum $||\nabla f(X)||_p$ because there are endless $X$, and a common practice is to approximate it by sampling \cite{iclr18evaluate}.
We use the training set to calculate $L_c$ because the model fits the training set better. For each full text in the whole training set, we first generate one rationale $Z$ with the generator and then calculate $\nabla f(Z)$. Note that here $\nabla f(Z) \in \mathcal{R}^{b\times d}$ is a matrix, where $b$ is the length of the rationale $Z$ which may varies across different rationales and $d$ is the dimension of the word vector. Corresponding to the distance metric of Definition~\ref{def:average distance}, we take the average along the length of $Z$ and get the unified $\nabla f(Z)\in \mathcal{R}^d$. Then we calculate $||\nabla f(Z)||_p$ and take the maximum value over the entire training set as the approximate $L_c$. It is worth noting that although different distance metrics correspond to different calculation methods of $L_c$, the theoretical derivation in Section~\ref{sec:lipcausedegeneration} doesn't involve any specific distance metric. In fact, the theoretical conclusions hold for any $d\in\mathcal{M}$. And we believe Definition~\ref{def:average distance} is somehow reasonable and will not undermine empirical qualitative conclusions in Section~\ref{sec:relation on lip and lr}.

\section{More results}

\subsection{The details of skewed predictor}\label{app:skewpred}
The experiment was first designed by \cite{interlocking}. It deliberately induces degeneration to verify the robustness of A2R compared to RNP. We first pretrain the predictor separately using only the first sentence of input text, and further cooperatively train the predictor initialized with the pretrained parameters and the generator randomly initialized using normal input text.
In Beer Reviews, the first sentence is usually about appearance. So, the predictor will overfit to the aspect of \emph{Appearance}, which is uninformative for \emph{Aroma} and \emph{Palate}. “\emph{skew}$k$” denotes the predictor is pre-trained for $k$ epochs. To make a fair comparison, we keep the pre-training process the same as that of A2R: batch-size=500 and learning-rate=0.001. 

\begin{table*}[t]
\caption{Some failure cases of DR. The \underline{underlined} piece of the text is the human-annotated rationale. Pieces of the text in \textcolor{blue}{blue} represent the rationales from DR. }
\begin{tabularx}{\textwidth}{X}
	\toprule
 \hline
      \textbf{Hotel-Cleanliness}\\
\textbf{Label:} positive, \textbf{Prediction:} positive\\
\textbf{Input text:}  my husband and i just spent two nights at the grand hotel francais , and we could not have been happier with our choice . in many ways , \underline{the hotel has exceeded our expectation} : the price was within our budget , \textcolor{blue}{breakfast was included , and the staff was friendly} , helpful and fluent in english . as other travelers have mentioned , the hotel is close to the nation metro station , which makes it easy to get around . the room size was just enough to fit two people , but \underline{we had a comfortable stay throughout} . overall , the hotel lives up to its high trip advisor rating . we would love to stay here again anytime .\\
\hline\hline
 \textbf{Beer-Aroma}\\
 \textbf{Label:} positive,  \textbf{Prediction:} positive\\
\textbf{Input text:} a- amber gold with a solid two maybe even three finger head . looks absolutely delicious , i dare say \textcolor{blue}{it is one of the best looking beers} i 've had . \underline{s- light citrus and hops . not a very strong {aroma}} t-wow , {the} hops , \textcolor{blue}{citrus} and pine blow out the taste buds , very tangy in taste , yet perfectly balanced , leaving a crisp dry taste to the palate . m-light and crisp feel with a nice tanginess thrown in the mix . d- could drink this all night , too bad i only have one more of this brew . notes : one of the best balanced and best tasting ipa 's i 've had to date . ipa fans you have to try this one .\\
\hline\hline

\end{tabularx}
\label{tab:fail}
\end{table*}

\begin{figure*}[t]
    \centering
    \subfigure[]{
    \includegraphics[width=0.48\columnwidth]{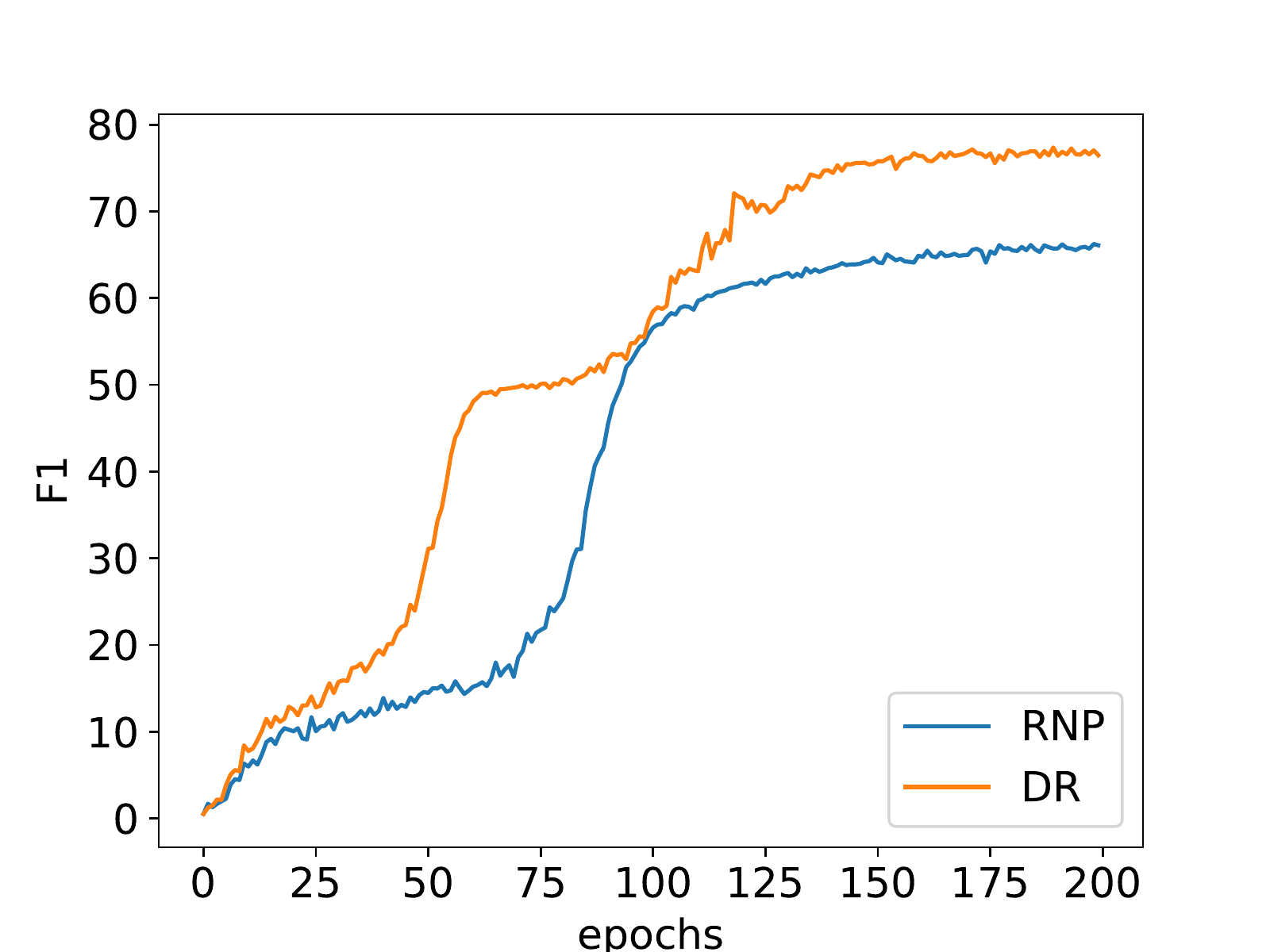}
        }
      \subfigure[]{
        \includegraphics[width=0.48\columnwidth]{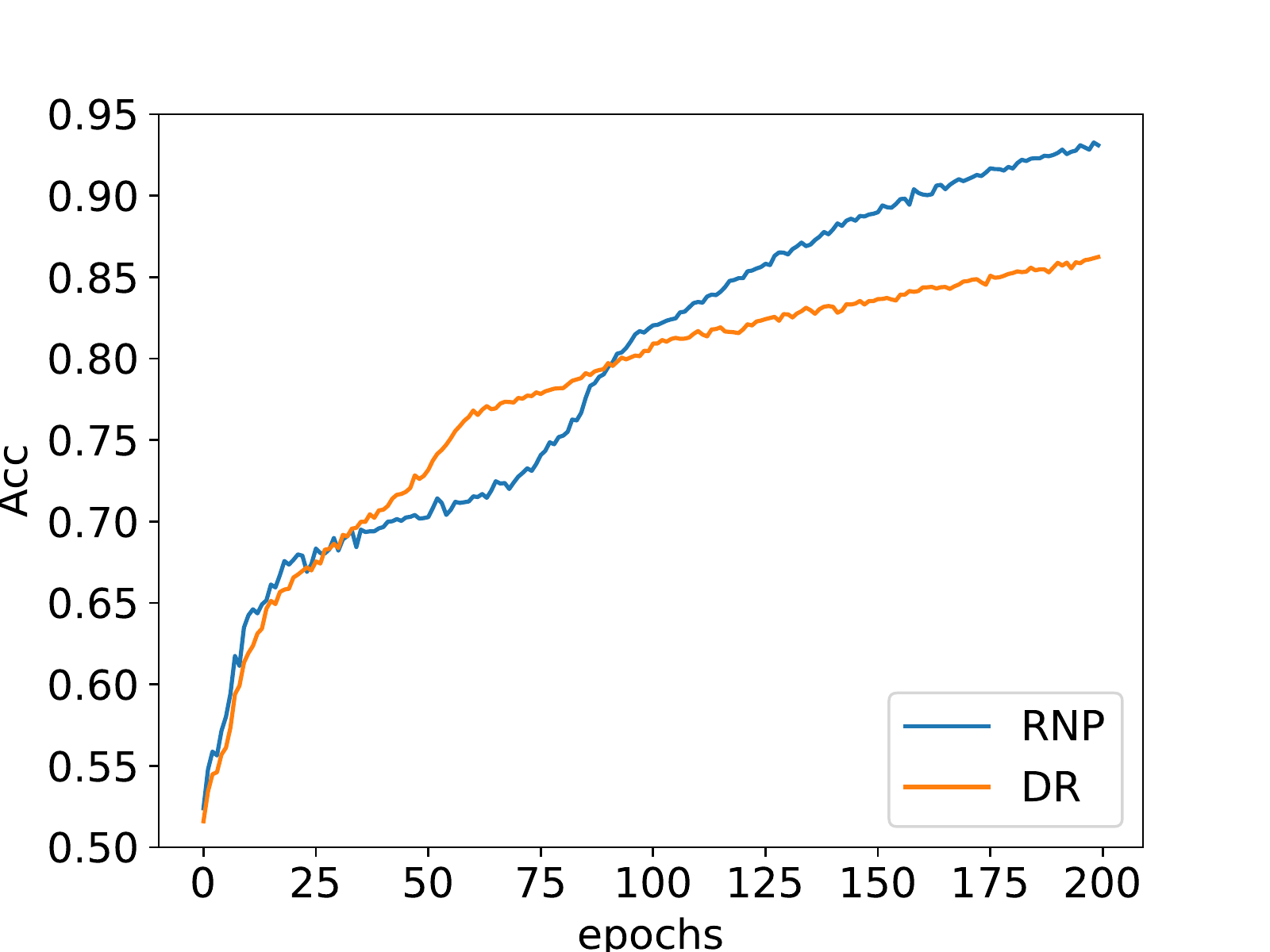}
        }
        \subfigure[]{
    \includegraphics[width=0.48\columnwidth]{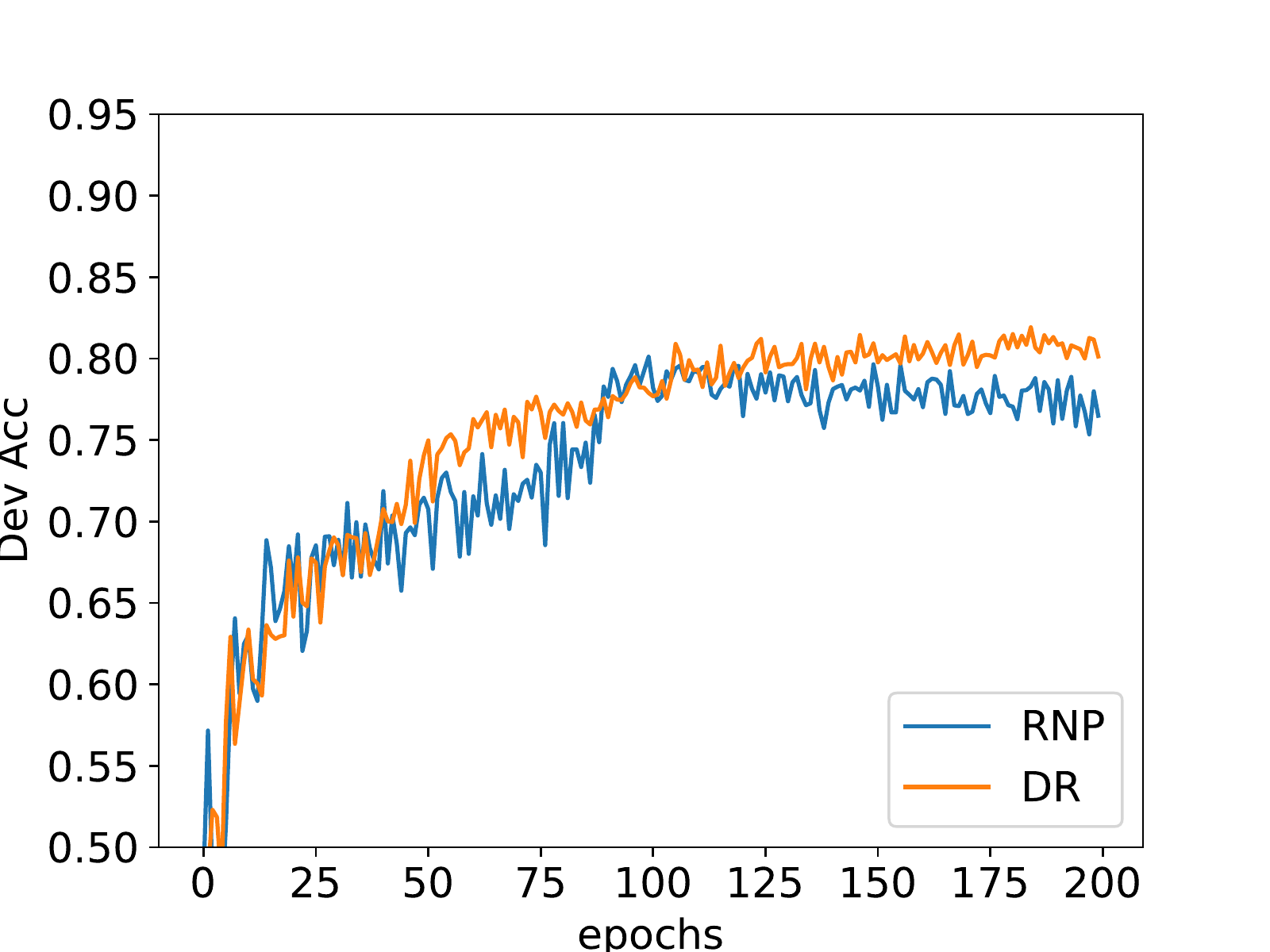}
        }
    \subfigure[]{
        \includegraphics[width=0.48\columnwidth]{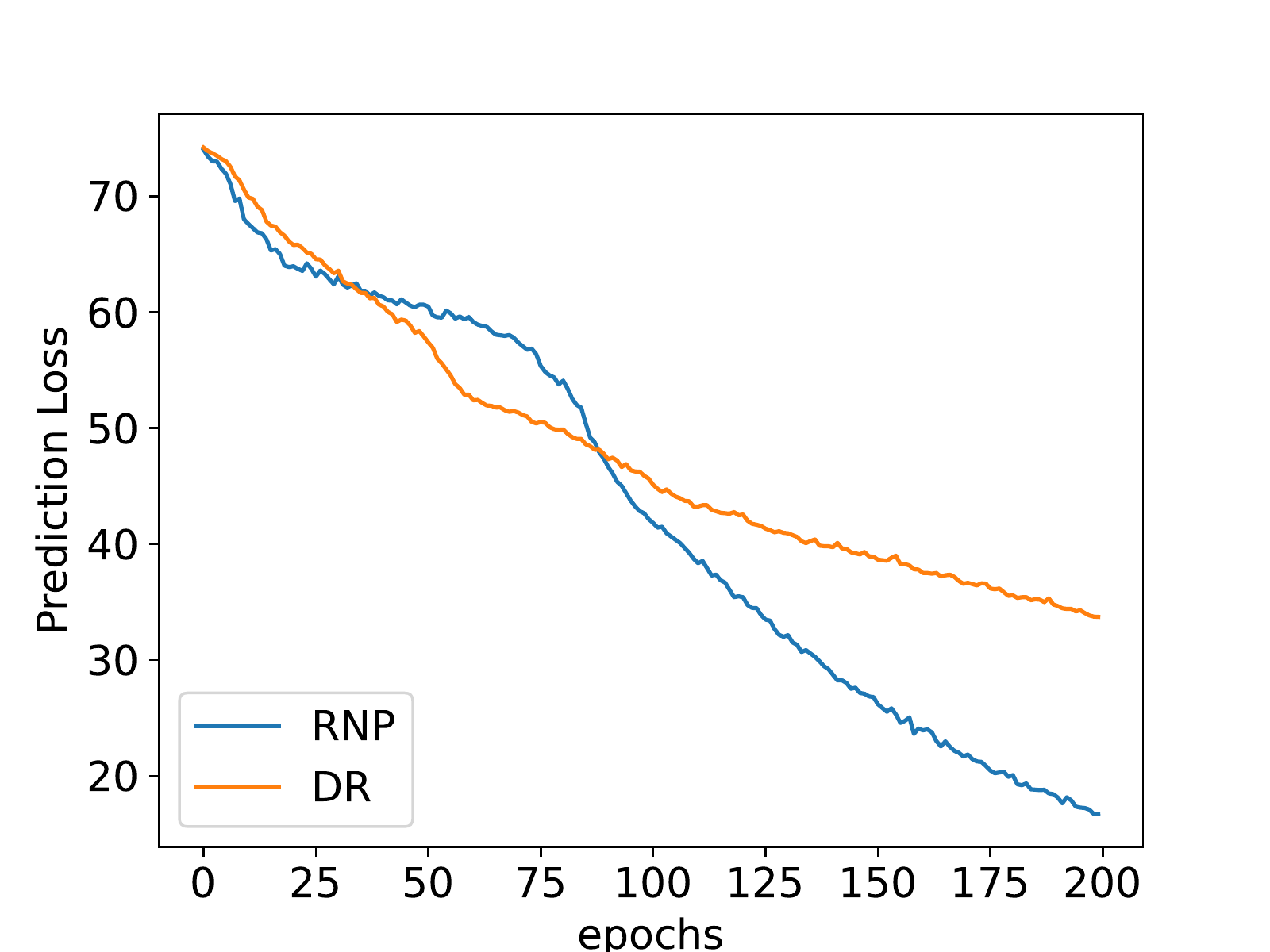}
    }
   \subfigure[]{
    \includegraphics[width=0.48\columnwidth]{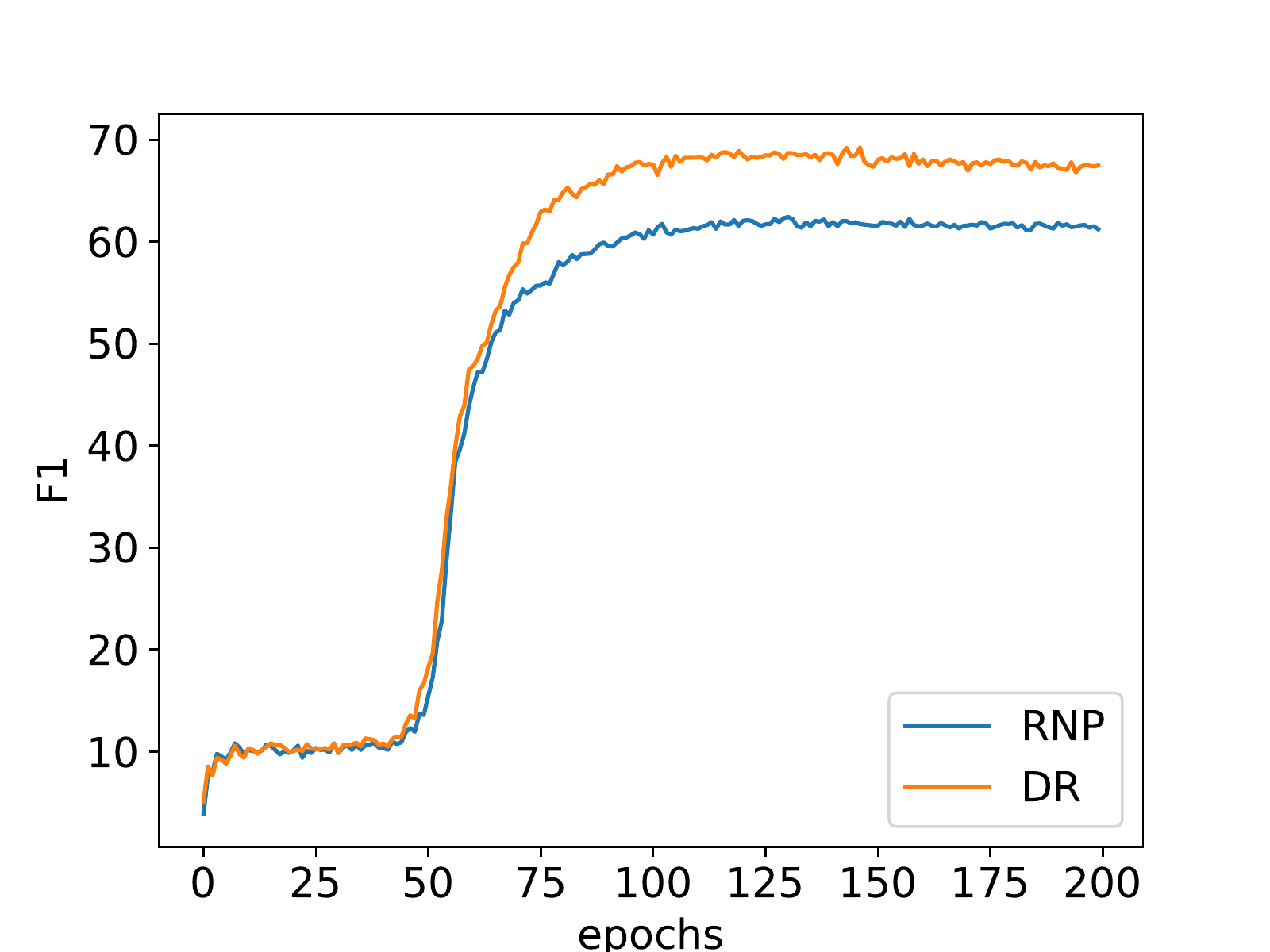}
        }
      \subfigure[]{
        \includegraphics[width=0.48\columnwidth]{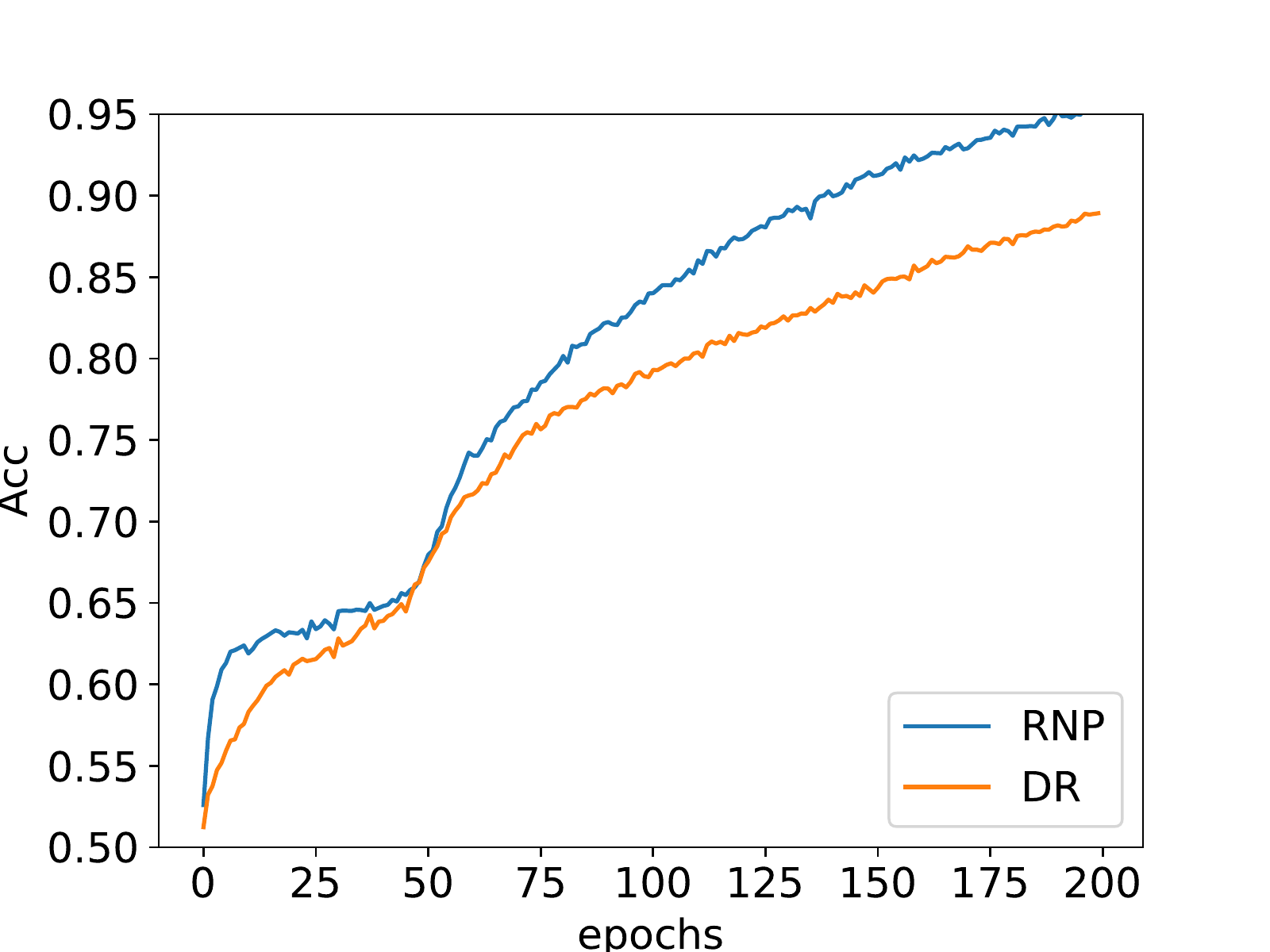}
        }
        \subfigure[]{
    \includegraphics[width=0.48\columnwidth]{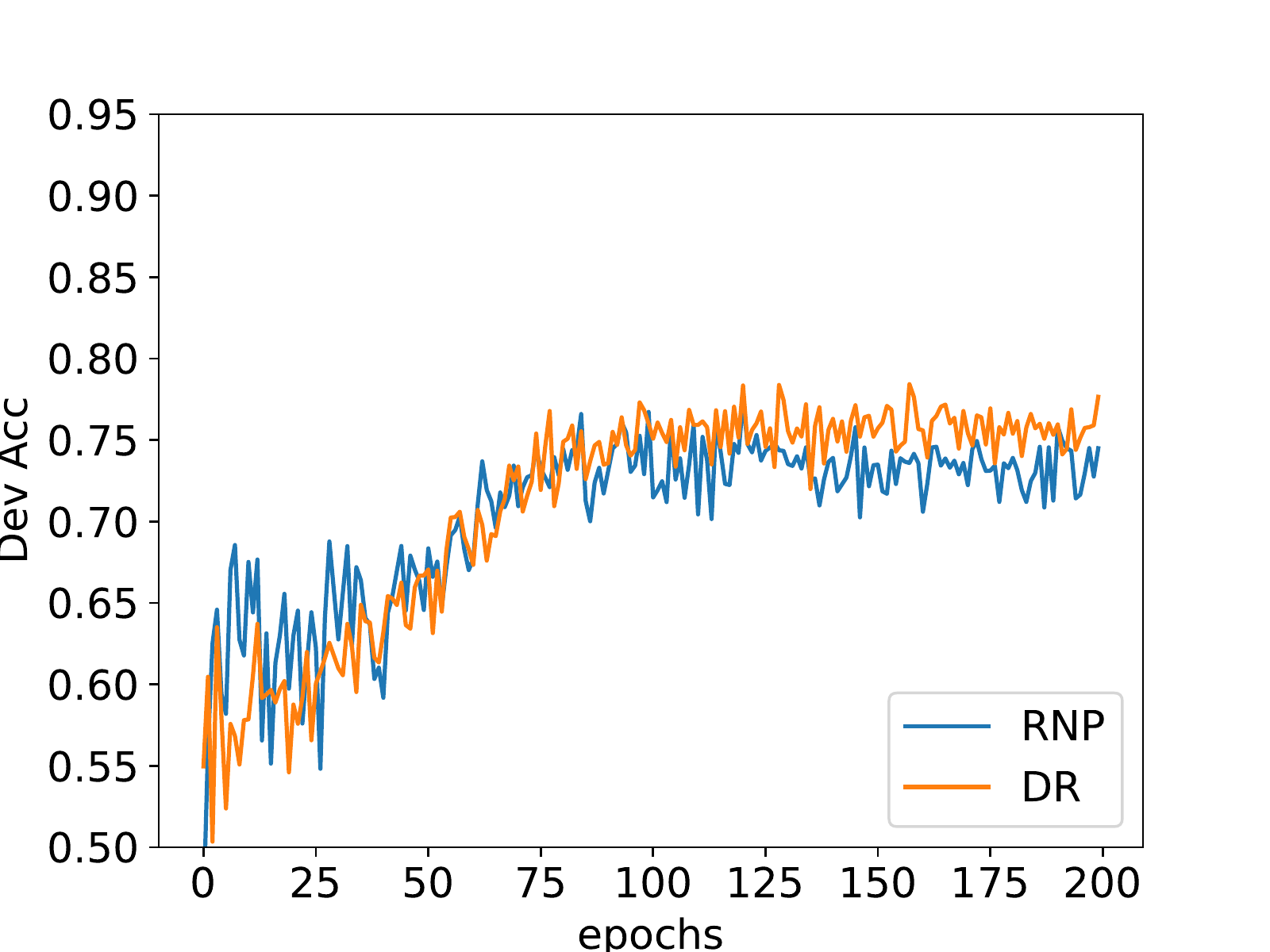}
        }
    \subfigure[]{
        \includegraphics[width=0.48\columnwidth]{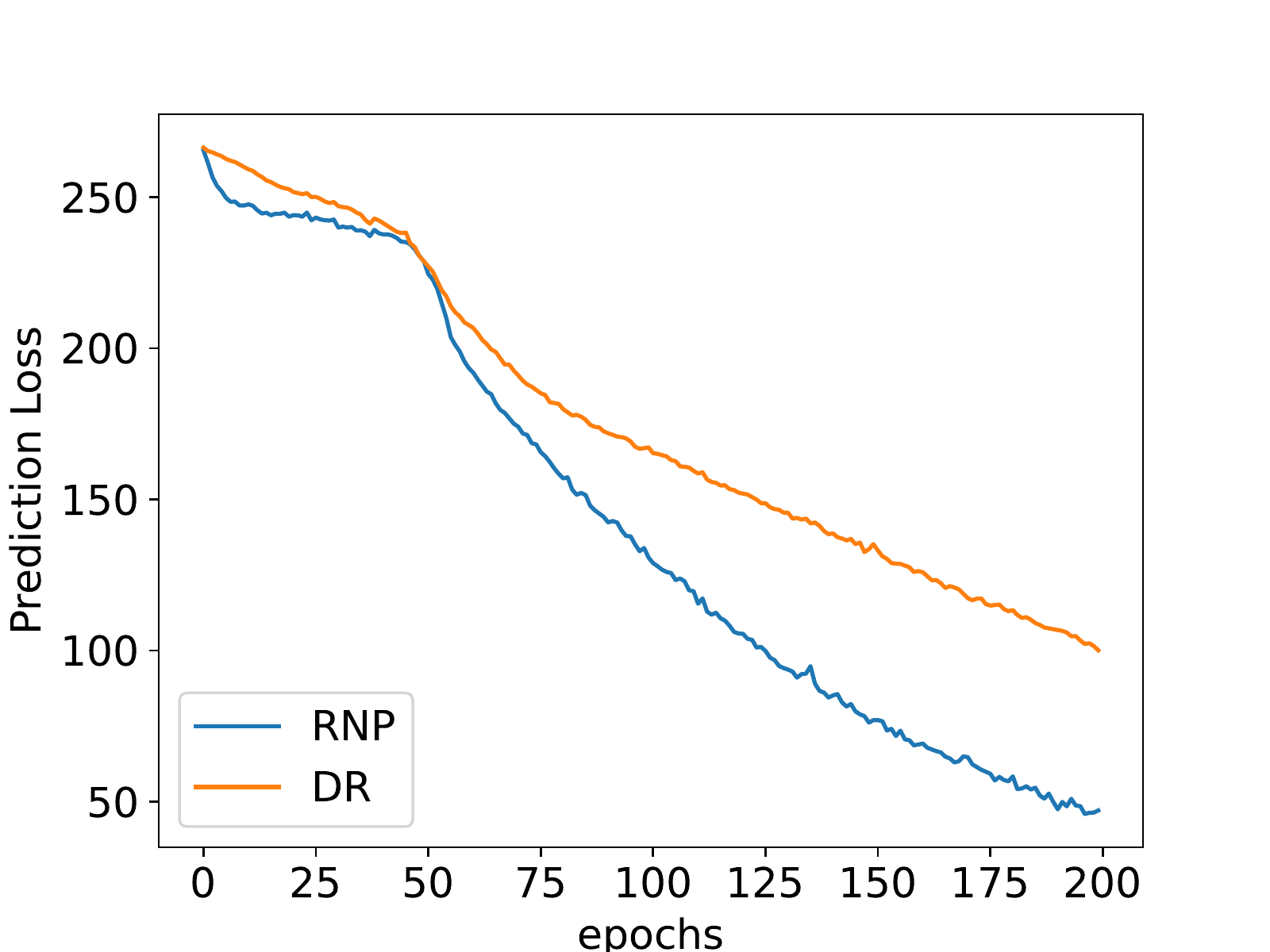}
    }
  \caption{The training process on \emph{Beer-Aroma} (a,b,c,d) and \emph{Beer-Palate} (e,f,g,h).}
  \label{fig:time_more}
\end{figure*}

\subsection{Failure cases}\label{app:fail}
To further pave ways to gain better insights of the model and help the line of rationalization research, we also provide some failure cases of our DR in Table~\ref{tab:fail}. We find that the model fails to make logical reasoning. For example, the ground truth of hotel cleanliness should be inferred from the word $``$comfortable$"$. However, our model makes judgments by selecting the rationales such as $``$breakfast $\cdots$ friendly$"$, which have little relevance to the cleanliness but are positive. We leave it as the future work.

\section{Theorem Proof}

\subsection{Proof of Theorem~\ref{the:rationale and lipschitz}}\label{proof:rationale and lipschitz}
We first copy Equation~\ref{eqa: distance larger than lip} here:
\begin{equation}
    d(Z_i,Z_j)\geq \frac{1-\epsilon_j-\epsilon_i}{L_c}. 
\end{equation}
And since we have $L_c\leq \frac{1-\epsilon_j-\epsilon_i}{\delta_d}$, we then get 
\begin{equation}
    d(Z_i,Z_j)\geq \delta_d,
\end{equation}
which is just the left part of Equation~\ref{eqa:assumption1}.
According to Assumption~\ref{assumption:distance threshold}, we then have 
\begin{equation}
    \Tilde{p}(Z_i,Z_j)\leq P_{te},
\end{equation}
where $\Tilde{p}(Z_i,Z_j)$ denotes the probability that $Z_i$ and $Z_j$ are not both informative.
Denoting ${p}(Z_i,Z_j)$ as the probability that $Z_i$ and $Z_j$ are both informative, we then have 
\begin{equation}
    {p}(Z_i,Z_j)\geq 1-P_{te}.
\end{equation}
The proof of Theorem~\ref{the:rationale and lipschitz} is completed.

\end{document}